\documentclass[letter,journal,compsoc]{IEEEtran}
\usepackage{cite}
\usepackage{soul}
\usepackage{xpatch}
\ifCLASSINFOpdf
\usepackage[pdftex]{graphicx}
\graphicspath{{../pdf/}{../jpeg/}}
\DeclareGraphicsExtensions{.pdf,.jpeg,.png}
\else
\usepackage[dvips]{graphicx}
\graphicspath{{../eps/}}
\DeclareGraphicsExtensions{.eps}
\fi
\usepackage{amsmath}
\makeatletter  
\newif\if@restonecol  
\makeatother

\usepackage[linesnumbered,ruled,vlined]{algorithm2e}
\usepackage{algpseudocode}  
\usepackage{ragged2e}
\ifCLASSOPTIONcompsoc
\usepackage{subfig}
\usepackage{color}
\usepackage{fixltx2e}
\usepackage{multirow}
\usepackage{color}
\usepackage{float}
\usepackage{pifont}
\usepackage{amsfonts}
\usepackage{booktabs}
\usepackage[normalem]{ulem} 
\usepackage{colortbl}
\usepackage[dvipsnames, svgnames, x11names]{xcolor}% 颜色支持
\usepackage[colorlinks=true,linkcolor=red,citecolor=green,urlcolor=magenta]{hyperref}
\begin{document}

\title{GeoThreat: Transferable Targeted Adversarial Attacks on Large Vision-Language Models for Remote Sensing Image Interpretation} 

\author{Yimin Fu, Yuefeng Bai, Baicheng Pan, Zhunga Liu, Michael K. Ng
\thanks{
\textit{(Corresponding authors: Michael K. Ng; Zhunga Liu.)}

The first two authors contributed equally to this work.

Yimin Fu, Baicheng Pan, and Michael K. Ng are with the Department of Mathematics, Hong Kong Baptist University, Hong Kong, China (e-mail: fuyimin@hkbu.edu.hk; baichengpan@life.hkbu.edu.hk; michael-ng@hkbu.edu.hk).

Yuefeng Bai and Zhunga Liu are with the School of Automation, Northwestern Polytechnical University, Xi'an, 710072, China (e-mail: baiyuefeng@mail.nwpu.edu.cn; liuzhunga@nwpu.edu.cn).
}}

\markboth{Manuscript Under Review}%
{Shell \MakeLowercase{\textit{et al.}}: A Sample Article Using IEEEtran.cls for IEEE Journals}

\IEEEtitleabstractindextext{%	
\begin{abstract}
\justifying Adversarial attacks against large vision-language models (LVLMs) serve as an effective means of assessing their robustness in cross-modal semantic understanding.
Existing studies mainly focus on corrupting visual inputs to induce predefined erroneous responses in general vision-language tasks, whereas corresponding investigations in remote sensing fields remain largely underexplored.
Compared with natural image understanding, remote sensing image interpretation requires joint reasoning over local discriminative cues and global scene context.
This poses additional challenges to achieving transferable semantic manipulation toward specified responses under black-box settings.
To tackle these challenges, we propose GeoThreat, a transferable targeted adversarial attack method against LVLMs for remote sensing image interpretation.
Specifically, GeoThreat modulates adversarial representations in accordance with the target content at both conceptual and perceptual levels.
The class tokens from surrogate image encoders are employed as conceptual representations, while perceptual representations are distilled from patch tokens of the adversarial example through collaborative importance estimation. 
Beyond merely rolling out attention scores across layers, we incorporate adversarial-target similarity gradients to more faithfully characterize the relevance of local visual cues to the intended semantic manipulation.
The perceptual representations are then dynamically aligned with target patch tokens in a cross-attentive manner, facilitating the adaptation of local cues toward designated semantic details.
Finally, adversarial perturbations are iteratively updated via ensemble-based joint optimization of conceptual calibration and perceptual adaptation.
Extensive experiments across diverse LVLMs demonstrate the superiority of GeoThreat in both transferability and controllability.
The code will be released at \url{https://github.com/fuyimin96/GeoThreat} upon acceptance.
\end{abstract}  

}

\maketitle

\IEEEdisplaynontitleabstractindextext

\IEEEpeerreviewmaketitle

\ifCLASSOPTIONcompsoc
\IEEEraisesectionheading{\section{Introduction}\label{sec:introduction}}
\else
\section{Introduction}
\label{introduction}
\fi
\IEEEPARstart{T}{he} exceptional multimodal understanding and reasoning capabilities of large vision-language models~(LVLMs)~\cite{liu2023visual,dai2023instructblip,achiam2023gpt,wang2024qwen2,li2025mini} have fostered their widespread adoption across diverse applications. 
Alongside this growing adoption, their vulnerability to adversarial perturbations has also raised substantial safety concerns~\cite{yuan2019adversarial,liu2024safety}.
To support reliable deployment in practice, adversarial attacks against LVLMs~\cite{liu2025survey} have been increasingly investigated as an effective approach for assessing model robustness and revealing potential weaknesses.
Given the imperceptibility of visual perturbations and their disruptive effect on cross-modal alignment~\cite{cui2024robustness}, existing methods typically craft adversarial examples based on input images to mislead LVLMs into generating erroneous responses.

\begin{figure}[]
\centering \vspace{-0.3cm}
\includegraphics[width=0.5\textwidth]{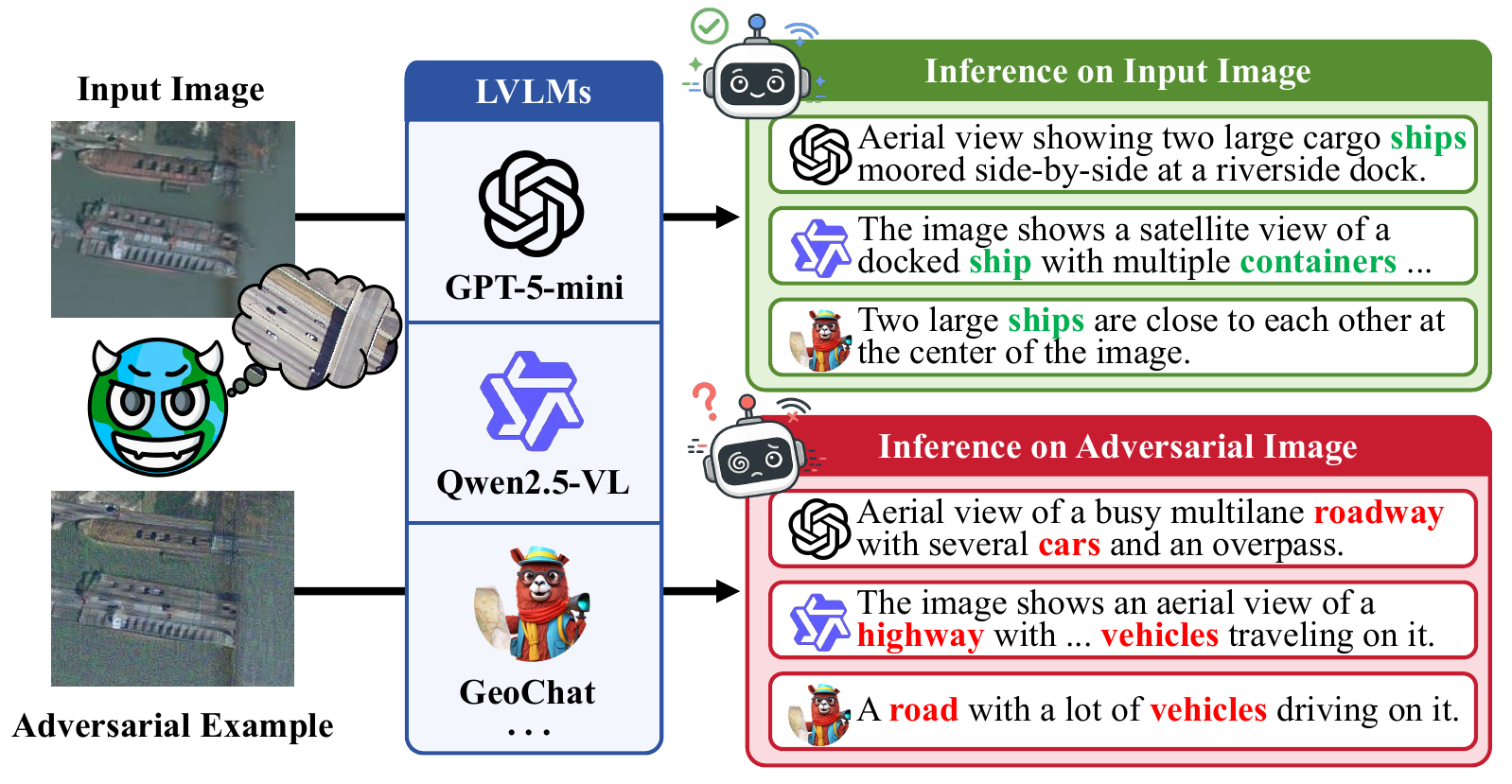}
\caption{Illustration of GeoThreat for transferable targeted adversarial attacks against LVLMs in remote sensing image interpretation.}
\label{rs_attack} 
\end{figure}

Since the model parameters and internal implementations of LVLMs are often inaccessible in practice, especially for commercial proprietary models, adversarial attacks against LVLMs are typically formulated under black-box settings~\cite{liu2017delving,zheng2025blackboxbench}.
Depending on the task objective, adversarial attacks against LVLMs can be further categorized into untargeted~\cite{yin2023vlattack} and targeted~\cite{zhao2023mf-ii} forms.
Beyond solely inducing erroneous responses, targeted attacks further require controllable semantic manipulation toward specified intents, which is regarded as the most realistic and challenging attack scenario.
Existing methods typically enhance the transferability and controllability of such targeted semantic manipulation by matching representations between adversarial examples and target samples, with pretrained CLIP~\cite{radford2021learning} image encoders commonly adopted as surrogates.
To further enhance adversarial transferability, model ensemble~\cite{chen2024diffusion,li2025mattack,jia2025foa} techniques are commonly employed to effectively exploit shared vulnerabilities across models, with perturbation updates derived from aggregated gradients over multiple surrogates.

Despite continued progress in misleading LVLMs toward specified responses under black-box settings, current research has predominantly focused on general vision-language tasks~\cite{zhang2024vision}.
With the increasing adoption of LVLMs in remote sensing~\cite{kuckreja2024geochat,irvin2025teochat,weng2025vision,chai2026like}, ensuring reasoning over remote sensing images has become a pressing requirement, especially in security-critical scenarios such as disaster assessment~\cite{chen2026integration} and urban monitoring~\cite{lin2025fedrsclip}.
However, research on targeted adversarial attacks against LVLMs for remote sensing image interpretation remains underexplored.
Compared with natural image understanding, the accurate interpretation of remote sensing images requires joint reasoning over local discriminative cues and global scene context~\cite{li2019deep}, which introduces distinct challenges for achieving transferable semantic manipulation toward specified responses.
In terms of transferability, the joint global-local reasoning paradigm exhibits substantial inter-model discrepancies, making adversarial example generation more prone to overfitting to surrogate-specific information.
In addition, the ambiguous content relevance of local perceptual cues, together with their complex dependencies on global scene context, further complicates the realization of controllable semantic manipulation toward specified responses.
Therefore, in this case, the generation of adversarial examples requires not only global conceptual matching with target samples, but also local perceptual alignment with the designated semantics.

Building upon this insight, we propose GeoThreat, a transferable targeted adversarial attack method against LVLMs for remote sensing image interpretation.
Specifically, GeoThreat addresses the transferability and controllability bottlenecks of attacks on LVLM-based remote sensing image interpretation through target-oriented representation modulation at both conceptual and perceptual levels.
During each iteration, the adversarial example and the target image are fed into the surrogate image encoder, where the extracted class tokens are employed to perform conceptual calibration.
Meanwhile, by integrating cross-layer attention with adversarial-target similarity, adversarial patch tokens that are relevant to the intended semantic manipulation are selectively identified as perceptual representations.
Then, the alignment between perceptual representations and target patch tokens is performed in a cross-attentive manner~\cite{vaswani2017attention}, thereby driving the adaptation of local perceptual cues toward the designated semantics.
Finally, adversarial example generation is guided by a joint optimization strategy of conceptual calibration and perceptual adaptation over model ensembles.
Extensive experiments across various LVLMs demonstrate that the proposed method outperforms state-of-the-art
attack methods.
As illustrated in Fig.~\ref{rs_attack}, GeoThreat successfully steers the interpretations of both general-purpose~\cite{achiam2023gpt,wang2024qwen2} and remote sensing-specific~\cite{kuckreja2024geochat} LVLMs toward the semantics of the target image under black-box settings.

Our main contribution can be summarized as follows:
\begin{enumerate}
\item We propose GeoThreat, a transferable targeted attack for LVLM-based remote sensing image interpretation, enabling controllable semantic manipulation through coordinated conceptual calibration and perceptual adaptation.
\item We develop a collaborative importance estimation approach to identify critical adversarial patch tokens and adapt their encoded perceptual cues toward the designated semantics in a cross-attentive manner.
\item We design a joint adversarial optimization strategy over ensemble surrogate models to effectively enhance the transferability and controllability of targeted adversarial examples against LVLMs.
\item We thoroughly benchmark adversarial attacks across general-purpose and remote sensing-specific LVLMs for remote sensing image interpretation.
The comprehensive robustness assessment reveals potential weaknesses of current LVLMs and provides a deeper understanding of their security risks in remote sensing tasks.
\end{enumerate}

The remainder of this paper is organized as follows. Section~\ref{sec2} reviews the related work. Section~\ref{sec3} introduces the problem formulation and technical details of the proposed method. Section~\ref{sec4} reports the experimental results and corresponding analyses. 
Finally, Section~\ref{sec5} concludes this paper and outlines potential directions for future research.

\section{Related Works}\label{sec2}
\subsection{Large Vision-Language Models}
Benefiting from their strong cross-modal understanding and reasoning capability, LVLMs have achieved substantial progress on diverse vision-language tasks, including image captioning~\cite{lu2025benchmarking}, visual question answering~\cite{Luu2024questioning}, and cross-modal retrieval~\cite{wang2024cross}. 
In terms of architecture, an LVLM~\cite{zhang2024vision} typically couples a pretrained vision encoder~\cite{radford2021learning,caron2021emerging} with a large language model (LLM)~\cite{chang2024survey} through a cross-modal alignment module, which projects visual patch embeddings into the input space of the LLM.
Li et al.~\cite{li2023blip} proposed BLIP-2, an efficient pre-training framework that employs a lightweight Querying Transformer to bridge the modality gap between image encoders and language models.
To enable general-purpose visual assistance, Liu et al.~\cite{liu2023visual} introduced LLaVA, which extends instruction tuning to multimodal scenarios with GPT-4-generated instruction data.
Zhu et al.~\cite{zhu2024minigpt} introduced MiniGPT-4, which aligns frozen visual features with an LLM through a lightweight projection layer to enable fluent vision-grounded language generation.
In addition to the aforementioned open-source LVLMs, closed-source commercial LVLMs, such as GPT-5~\cite{openai2025gpt5}, Claude~\cite{anthropic2025claude4}, and Gemini~\cite{gemini2023gemini}, have demonstrated advanced multimodal understanding and reasoning capabilities, benefiting from large-scale training data and system-level optimization.

In addition to general vision-language applications, growing research efforts have also been devoted to advancing the adoption of LVLMs in the remote sensing domain.
As an early exploration of remote sensing-specific vision-language foundation models, Liu et al.~\cite{liu2024remoteclip} proposed RemoteCLIP to learn semantically rich visual features.
Based on a high-quality remote sensing image captioning dataset, Hu et al.~\cite{hu2025rsgpt} systematically benchmarked VLMs for remote sensing vision-language understanding.
Kuckreja et al.~\cite{kuckreja2024geochat} developed GeoChat, a multitask conversational assistant for remote sensing image interpretation with strong spatial grounding and reasoning capabilities.
To compensate for the insufficient consideration of geospatial variability, Muhtar et al.~\cite{muhtar2024lhrsbot} leveraged volunteered geographic information to empower remote sensing image understanding of LVLMs.
Zhan et al.~\cite{zhan2025skyeyegpt} unified multi-granularity remote sensing vision-language tasks through large-scale instruction data and a two-stage fine-tuning strategy.
Zhang et al.~\cite{zhang2024earthgpt} designed EarthGPT, which unifies diverse remote sensing interpretation tasks across multi-sensor imagery.
For temporal earth observation data, Irvin et al.~\cite{irvin2025teochat} proposed TEOChat, which leverages image references in prompts to enable conversational understanding of remote sensing image sequences.

\subsection{Adversarial Attacks on LVLMs}
Conventional adversarial attacks mainly aim to induce incorrect predictions in image classification tasks under white-box~\cite{kurakin2016adversarial,kurakin2017adversarial,madry2018towards} or black-box~\cite{dong2018boosting,zheng2025blackboxbench} settings, depending on the accessibility of the victim model.
In comparison, adversarial attacks against LVLMs are often formulated under a targeted paradigm, aiming to steer model responses toward predefined erroneous content.
Given that LVLMs are often deployed as proprietary services in practice, such semantic manipulation is typically achieved by crafting transferable adversarial examples on surrogate vision encoders.

Zhao et al.~\cite{zhao2023mf-ii} developed AttackVLM to craft transferable visual perturbations against LVLMs by matching image or text embeddings in pretrained vision-language models under black-box settings.
For further adversarial transferability improvement, Dong et al.~\cite{dong2023robust} incorporated frequency-domain transformations and model ensembling to exploit common weaknesses across model architectures.
Guo et al.~\cite{guo2024efficient} introduced AdvDiffVLM, a diffusion-based attack framework that embeds adversarial objectives into the denoising process to generate transferable visual perturbations against LVLMs.
To overcome the scalability limitations caused by the reliance on target labels, Zhang et al.~\cite{zhang2025anyattack} proposed AnyAttack, a self-supervised attack framework through large-scale pre-training.
Beyond the output perspective, increasing attention has been paid to steering LVLM responses toward designated semantics by exploiting intermediate representations of vision encoders.
Li et al.~\cite{li2025mattack} induced desired misinterpretations in closed-source LVLMs by aligning locally cropped adversarial images with target images in the embedding space.
To improve generalization across diverse tasks, Mei et al.~\cite{mei2026veattack} proposed the vision encoder attack (VEAttack), which generates downstream-agnostic perturbations through token-level disruption of image features.
Jia et al.~\cite{jia2025foa} proposed FOA-Attack, which reformulates local feature alignment as an optimal transport problem to alleviate overfitting to surrogate-specific characteristics.
Instead of directly using entangled patch features, Nie et al.~\cite{nie2026vattack} improved the controllability of semantic shifts in LVLMs by targeting disentangled value features.
Li et al.~\cite{li2026mpcattack} proposed a multi-paradigm collaborative attack (MPCAttack) approach to jointly optimize aggregated visual-textual features, thereby mitigating the transferability limitation caused by insufficient representation diversity.

\subsection{Adversarial Machine Learning in Remote Sensing}
Beyond general computer vision and multimodal tasks, adversarial machine learning has also evolved into an ongoing attack-defense race in the field of remote sensing to ensure security and trustworthiness.

On the attack side, Xu et al.~\cite{xu2020assessing} first provided a systematic assessment of adversarial threats to deep neural networks (DNNs) for remote sensing scene classification, covering both targeted and untargeted attack scenarios.
In addition, Chen et al.~\cite{chen2021empirical} conducted an empirical study to further reveal the vulnerability of DNN-based remote sensing scene classification models to adversarial examples.
By incorporating intermediate-layer representation mixing into perturbation generation, Xu et al.~\cite{xu2022universal} proposed a mixup attack to enhance adversarial transferability against remote sensing scene classifiers.
Fu et al.~\cite{fu2024sfcot} integrated transformations in both the spatial and frequency domains to enrich the simulation of target-related input patterns, thereby further enhancing the effectiveness of black-box attacks.
In addition to image classification, Sun et al.~\cite{sun2024task} enhanced attack effectiveness in optical remote sensing object detection through importance-aware candidate selection. 
Based on a mask-constrained perturbation generation strategy, Bai et al.~\cite{bai2024stealthy} realized a stealthy attack on specific victim classes for remote sensing semantic segmentation.
Recently, Fu et al.~\cite{fu2026transferability} reinforced adversarial transferability in remote sensing scene classification through a coordinated action-execution pipeline for transformation composition.
For other remote sensing modalities, Shi et al.~\cite{shi2021hyperspectral} investigated the impact of adversarial examples on hyperspectral image (HSI) interpretation, showing that spectral-spatial representations can be substantially disrupted.
Li et al.~\cite{li2025sparse} further improved the efficiency of attacks for HSI classification by leveraging sparse unmixing to generate physically consistent class-specific perturbations.
Based on the unique imaging mechanism of synthetic aperture radar (SAR), Peng et al.~\cite{peng2022scattering} proposed a scattering model-guided adversarial attack for SAR target recognition.

In parallel with the attack side, increasing efforts have also been devoted to defense strategies for improving adversarial robustness in remote sensing.
Su et al.~\cite{su2023reconstruction} proposed a reconstruction-assisted adversarial training framework to mitigate the interference of adversarial perturbations on remote sensing scene classification.
Benefiting from the strong denoising capability of diffusion models, Yu et al.~\cite{yu2024diffusion} developed a universal defense framework to restore adversarial examples to underlying data distributions.
Qi et al.~\cite{qi2024masked} integrated self-supervised spectral reconstruction and dynamic graph embedding to mitigate the adverse influence of adversarial perturbations.
Pathak et al.~\cite{pathak2024model} improved the reliability of drone-based object detection by formulating the defense against patch attacks as a recovery task for adversarially occluded object regions.
To defend against adversarial attacks in HSI classification, Xu et al.~\cite{xu2021sacnet} exploited global contextual information to enhance the adversarial robustness of DNNs.
Guided by the scattering priors, Liu et al.~\cite{liu2026scattering} employed a bi-directional information bottleneck to remove class-irrelevant information and suppress adversarial interference in SAR target recognition.

\begin{figure*}[]
\centering 
\includegraphics[width=1\textwidth]{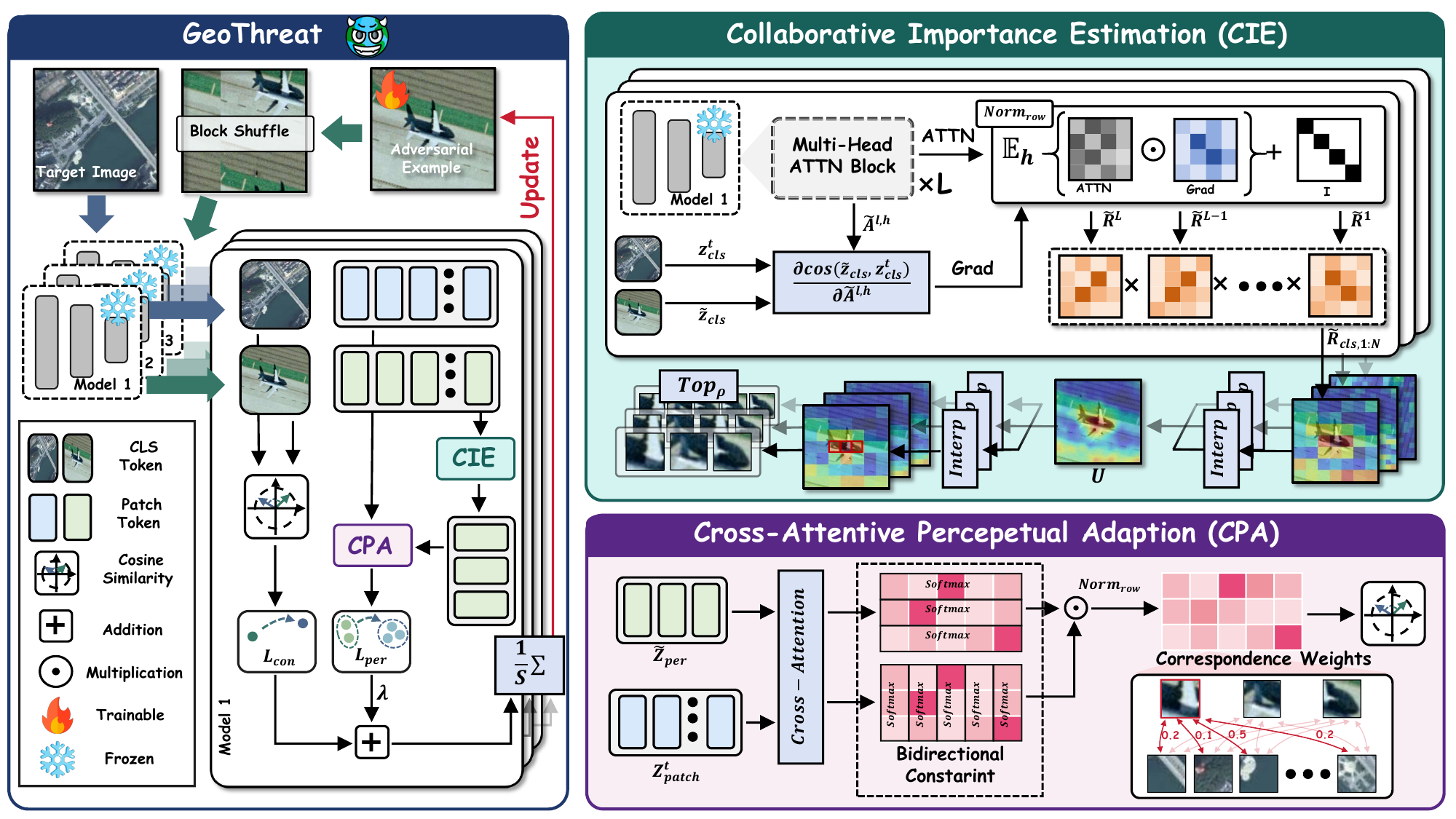}
\caption{Illustration of GeoThreat for transferable targeted adversarial attacks against LVLMs in remote sensing image interpretation. The proposed attack pipeline comprises three key components: collaborative importance estimation, cross-attentive perceptual adaptation, and ensemble-based joint adversarial optimization.}
\label{pipeline} 
\end{figure*}

\section{Methodology}\label{sec3}
The pipeline of the proposed GeoThreat for generating transferable targeted adversarial examples against LVLMs is illustrated in Fig.~\ref{pipeline}.
The randomly block-shuffled adversarial example, together with its corresponding target image, is first fed into the pretrained vision encoder.
Then, critical patch tokens of the adversarial example are identified through a collaborative importance estimation approach.
Finally, guided by the joint optimization of conceptual calibration and perceptual adaptation, the adversarial example is iteratively updated over the surrogate ensemble.
In this section, we first introduce the problem formulation of transferable targeted attacks for LVLM-based remote sensing image interpretation. Then, we present the technical details of GeoThreat for improving the transferability and controllability of targeted semantic manipulation.

\subsection{Problem Formulation}
Given a remote sensing image $x$ and a textual query $q$, remote sensing image interpretation with an LVLM $g_{\phi}(\cdot)$ can be represented as generating a response $y$ conditioned on the multimodal input pair:
\begin{equation}
y=g_{\phi}(x,q),
\end{equation}
where $\phi$ denotes the parameters of the LVLM.
Under the targeted attack scenario, the objective of adversarial attacks on LVLMs is to steer the model outputs toward designated semantics of the target image $x_t$.
Since the parameters $\phi$ of the victim LVLM are often inaccessible in practice, directly performing such semantic manipulation based on output-level information is infeasible.
In this case, existing methods typically employ a pretrained vision encoder $f_{\theta}(\cdot)$ as a surrogate to realize transfer-based attacks.
Specifically, the attack process is formulated as a constrained optimization problem, where visual perturbations $\delta$ are iteratively updated by aligning the representations of the adversarial image $\tilde{x}=x+\delta$ and the target image $x_t$:
\begin{equation}
\mathop{\min}\limits_{\lVert \delta\rVert_{\infty}\leq\epsilon}
\mathcal{L}
\left(
f_{\theta}(\tilde{x}),
f_{\theta}(x_t)
\right),
\end{equation}
where $\epsilon$ denotes the maximum perturbation magnitude under $\ell_{\infty}$-norm constraint.
The loss function $\mathcal{L}(\cdot,\cdot)$ is used to measure the representation discrepancy.
To further enhance the transferability, an ensemble of surrogate vision encoders $\mathcal{F}=\{f_{\theta}^s\}_{s=1}^{S}$ with size $S$ is commonly adopted to provide aggregated guidance for adversarial example generation:
\begin{equation}
\mathcal{L}_{ens}(\tilde{x},x_t)=\frac{1}{|\mathcal{F}|} \sum_{f_\theta \in \mathcal{F}} \mathcal{L}\left(f_\theta(\tilde{x}), f_\theta\left(x_t\right)\right).
\end{equation}
Starting from $\tilde{x}^{0}=x$, the adversarial example at the $k$-th iteration is updated as:
\begin{equation}
\tilde{x}^{k}=\operatorname{Clip}_{x, \epsilon}(\tilde{x}^{k-1}-\alpha \cdot \operatorname{sign}(\nabla_{\tilde{x}^{k-1}} \mathcal{L}_{ens}(\tilde{x}^{k-1},x_t))),
\end{equation}
where $\mathrm{Clip}_{x,\epsilon}(\cdot)$ denotes the pixel-wise clipping operation, and $\alpha$ is the attack step size.
For clarity, the following derivation is presented within a single iteration. Unless otherwise specified, the iteration superscript is omitted, and the current adversarial example is denoted by $\tilde{x}$.

\subsection{Collaborative Importance Estimation}
Existing adversarial attack methods typically mislead black-box LVLMs into producing specified outputs by matching global visual representations between adversarial and target images.
Due to the complex spatial heterogeneity and contextual dependencies inherent in remote sensing images, their interpretation is commonly grounded in joint reasoning over local discriminative cues and global scene context.
In this case, optimization signals derived solely from global conceptual calibration are insufficient to support transferable targeted semantic manipulation, necessitating the incorporation of local discriminative cues into adversarial example generation.
A commonly adopted approach to identify critical image patches is to roll out attention scores across transformer layers.
However, such attention scores mainly reflect the decision relevance of patch tokens for the input image, rather than their responsiveness in driving the transition toward the target semantics.
Consequently, perturbation optimization may become overly biased toward surrogate-specific recognition patterns of the input image, thereby further limiting the effectiveness of transferable targeted attacks.

To overcome this limitation, we develop a collaborative importance estimation approach to identify adversarial patch tokens that are both decision-relevant and target-responsive.
First, the adversarial example $\tilde{x}$ and the target image $x_t$ are fed into a surrogate vision encoder $f_{\theta}(\cdot)$, e.g., a transformer-based CLIP image encoder, to extract token-wise visual representations:
\begin{equation}
\begin{aligned}
f_\theta(\tilde{x})& =\left[\tilde{\mathbf{z}}_{\text {cls}} ; \tilde{\mathbf{Z}}_{\text {patch}}\right]=\left[\tilde{\mathbf{z}}_{\text{cls }} ; \tilde{\mathbf{z}}_1, \ldots, \tilde{\mathbf{z}}_N\right], \\
f_\theta\left(x_t\right)& =\left[\mathbf{z}_{\text {cls}}^t ; \mathbf{Z}_{\text {patch}}^t\right]=\left[\mathbf{z}_{\text {cls}}^t ; \mathbf{z}_1^t, \ldots, \mathbf{z}_N^t\right],
\end{aligned}
\end{equation}
where $\tilde{\mathbf{z}}_{cls}$, $\mathbf{z}_{cls}^t \in\mathbb{R}^{d}$ denote the class tokens of the adversarial and target images, respectively. 
$\tilde{\mathbf{Z}}_{\text {patch}}=\{\tilde{\mathbf{z}}_{i}\}_{i=1}^N$, ${\mathbf{Z}}_{\text {patch}}^t=\{\mathbf{z}_{i}^t\}_{i=1}^N \in\mathbb{R}^{N\times d}$ denote their corresponding patch-token sequences with embedding dimension $d$.
Then, conceptual calibration is performed by maximizing the cosine similarity between the adversarial and target class tokens:
\begin{equation}
\mathcal{L}_{\mathrm{con}}=1-\cos \left(\tilde{\mathbf{z}}_{\mathrm{cls}}, \mathbf{z}_{\mathrm{cls}}^t\right)=1-\frac{\langle\tilde{\mathbf{z}}_{\mathrm{cls}}, \mathbf{z}_{\mathrm{cls}}^t\rangle}{\left\|\tilde{\mathbf{z}}_{\mathrm{cls}}\right\|_2\left\|\mathbf{z}_{\mathrm{cls}}^t\right\|_2},
\end{equation}
where $\langle \cdot,\cdot\rangle$ is the inner product, and $\lVert \cdot\rVert$ is the $\ell_{2}$-norm.
Meanwhile, the adversarial-target similarity is further leveraged to estimate the importance of adversarial patch tokens in driving the intended semantic manipulation.
During the forward pass of the adversarial example through the surrogate vision encoder, the self-attention matrix in the $h$-th head of the $l$-th layer is computed as:
\begin{equation}
\tilde{\mathbf{A}}^{l, h}=\operatorname{Softmax}\left(\frac{\tilde{\mathbf{Q}}^{l, h}\left(\tilde{\mathbf{K}}^{l, h}\right)^{\top}}{\sqrt{d_h}}\right) \in \mathbb{R}^{(N+1) \times(N+1)},
\end{equation}
where $\tilde{\mathbf{Q}}^{l, h}$ and $\tilde{\mathbf{K}}^{l, h}$ are the query and key matrices of the adversarial token sequence in the $h$-th attention head in the $l$-th layer, and $d_h$ denotes the dimension of each attention head.
Instead of relying solely on attention weights, we estimate patch importance at each layer by integrating attention weights with their corresponding conceptual-similarity sensitivities across $N_h$ attention heads:
\begin{equation}
\tilde{\mathbf{R}}^l=\operatorname{Norm}_{\mathrm{row}}\left(\mathbf{I}+\frac{1}{N_h} \sum_{h=1}^{N_h} \operatorname{ReLU}\left(\tilde{\mathbf{A}}^{l, h} \odot \nabla^{l, h}\right)\right),
\end{equation}
where the sensitivity term is defined as
\begin{equation}
\nabla^{l, h}=\frac{\partial \cos \left(\tilde{\mathbf{z}}_{\mathrm{cls}}, \mathbf{z}_{\mathrm{cls}}^t\right)}
{\partial \tilde{\mathbf{A}}^{l, h}}
\in \mathbb{R}^{(N+1)\times (N+1)}.
\end{equation}
Here, $\operatorname{ReLU}(\cdot)$, $\mathbf{I}$, $\operatorname{Norm}_\mathrm{row}(\cdot)$, and $\odot$ denote the ReLU activation function, identity matrix, row-wise normalization, and element-wise multiplication, respectively.
On this basis, the collaborative importance matrix is obtained through successive matrix multiplication across $L$ transformer layers:
\begin{equation}
\tilde{\mathbf{R}}
=\tilde{\mathbf{R}}^{L} \tilde{\mathbf{R}}^{L-1}
\cdots \tilde{\mathbf{R}}^{1} \in \mathbb{R}^{(N+1)\times (N+1)}.
\end{equation}
Finally, critical adversarial patch tokens are identified according to the class-to-patch scores in the collaborative importance matrix.
The indices of the top-$\rho$ patch tokens are selected as:
\begin{equation}\label{select_patch}
\mathcal{I}_\rho=\operatorname{Top}_\rho\left(\tilde{\mathbf{R}}_{\mathrm{cls}, 1: N}\right), \quad \rho \in(0,1],
\end{equation}
where $\tilde{\mathbf{R}}_{\mathrm{cls}, 1: N}$ represents the importance scores associated with the class-token row, with the self-relevance term excluded.
The selected patch tokens are employed as perceptual representations for subsequent adaptation of local cues:
\begin{equation}\label{get_patch}
\tilde{\mathbf{Z}}_{\text {per}}=\tilde{\mathbf{Z}}_{\text {patch}}[\mathcal{I}_\rho] \in\mathbb{R}^{N_{\rho}\times d},
\end{equation}
where $N_{\rho}={\lceil \rho N \rceil}$.

\subsection{Cross-Attentive Perceptual Adaptation}
Upon identifying critical perceptual representations, adapting local perceptual cues toward the designated semantics hinges on appropriate target correspondence modeling and representation alignment.
Given the spatial heterogeneity and complex semantic interdependencies in remote sensing images, the complete target patch-token sequence provides a more comprehensive semantic reference for establishing target-aware correspondences.
In addition, compared with a fixed alignment scheme, adaptive representation alignment conditioned on the collective dependencies between perceptual representations and target patch tokens enables more controllable semantic manipulation.

Motivated by this insight, we perform the adaptation of local perceptual cues toward the designated semantics in a cross-attentive manner.
Specifically, the perceptual representations $\tilde{\mathbf{Z}}_{\text {per}}$ and the target-image patch-token sequence $\mathbf{Z}_{\text {patch}}^t$ are used as queries and keys, respectively, to compute the cross-attentive patch-level affinity matrix
\begin{equation} \label{eq:get_cross_a}
\mathbf{M}
=\frac{\tilde{\mathbf{Z}}_{\mathrm{per}} \left(\mathbf{Z}_{\mathrm{patch}}^{t}\right)^{\top}
}{\sqrt{d}} \in \mathbb{R}^{N_{\rho} \times N}.
\end{equation}
Then, bidirectional correspondence distributions are derived from the affinity matrix by applying temperature-scaled softmax normalization along different dimensions:
\begin{equation} \label{eq:bidirection}
\begin{aligned}
\mathbf{P}_{ij}^{\tilde{x} \to x_t}
&=\frac{\exp \left( \mathbf{M}_{ij} / \tau \right)}{\sum_{j'=1}^{N} \exp \left( \mathbf{M}_{ij'} / \tau \right)}, \quad \mathbf{P}^{\tilde{x} \to x_t} \in \mathbb{R}^{ N_{\rho}  \times N}, \\
\mathbf{P}_{ji}^{x_t \to \tilde{x}}
&= \frac{\exp \left( \mathbf{M}_{ij} / \tau \right)
}{\sum_{i'=1}^{N_{\rho}} \exp \left( \mathbf{M}_{i'j} / \tau \right)},
\quad \mathbf{P}^{x_t \to \tilde{x}} \in \mathbb{R}^{N \times  N_{\rho} },
\end{aligned}
\end{equation}
where $\tau$ is the temperature parameter.
$\mathbf{A}^{\tilde{x} \to x_t}$ and $\mathbf{A}^{x_t \to \tilde{x}}$ quantify the correspondence weights from adversarial perceptual representations to target patch tokens and from target patch tokens to adversarial perceptual representations, respectively.
Afterward, the two directional distributions are integrated into mutual correspondence weights to more faithfully characterize the adversarial-target dependencies for adapting local cues toward the designated semantic details:
\begin{equation} \label{eq:get_weight}
\mathbf{P} = \mathrm{Norm}_{\mathrm{row}}
\left(\mathbf{P}^{\tilde{x} \to x_t}\odot
\left(\mathbf{P}^{x_t \to \tilde{x}}\right)^{\top}
\right) \in \mathbb{R}^{N_{\rho}\times N}.
\end{equation}
Based on the mutual correspondence weights, target patch tokens are aggregated to form query-conditioned semantic references:
\begin{equation} \label{eq:get_ref}
\mathbf{Z}_{\text {ref}}
=\mathbf{P}\mathbf{Z}_{\text {patch}}^t \in \mathbb{R}^{N_{\rho} \times d}.
\end{equation}
Finally, the adaptation of local perceptual cues is performed by adaptively aligning their representations with the corresponding target semantic references:
\begin{equation}
\begin{aligned}
\mathcal{L}_{\mathrm{per}}
&=1-\frac{1}{N_{\rho}}
\sum_{i=1}^{N_{\rho}}\cos
\left(\tilde{\mathbf{z}}_{\mathrm{per}, i}, \mathbf{z}_{\mathrm{ref}, i}\right)\\
&=1-\frac{1}{N_{\rho}}\sum_{i=1}^{N_{\rho}} \frac{\left\langle \tilde{\mathbf{z}}_{\mathrm{per}, i}, \mathbf{z}_{\mathrm{ref}, i} \right\rangle}{
\left\| \tilde{\mathbf{z}}_{\mathrm{per}, i} \right\|_2 \left\|\mathbf{z}_{\mathrm{ref}, i}
\right\|_2},
\end{aligned}
\end{equation}
where $\tilde{\mathbf{z}}_{\mathrm{per}, i} $ and $\mathbf{z}_{\mathrm{ref}, i}$ denote the $i$-th rows of $\tilde{\mathbf{Z}}_{\mathrm{per}}$ and $\mathbf{Z}_{\text {ref}}$, respectively.

\subsection{Joint Adversarial Optimization}
Without access to the internal information of victim LVLMs under black-box settings, a common transferability-enhancement strategy is to leverage an ensemble of surrogate models to provide more generalizable optimization signals for perturbation update.
In GeoThreat, beyond coupling global conceptual and local perceptual losses, heterogeneity among surrogate vision encoders also introduces inconsistency into the collaborative importance matrices.
Therefore, the coherent integration and harmonization of surrogate-derived information within the ensemble are crucial for further strengthening the transferability and controllability of the generated adversarial examples.

In pursuit of this objective, we design a joint adversarial optimization strategy over an ensemble of surrogate vision encoders to steer adversarial example generation toward the designated semantics.
Given an ensemble of $S$ surrogate vision encoders $\mathcal{F}=\{f_{\theta_s}\}_{s=1}^{S}$, the token-wise visual representations of the adversarial example $\tilde{x}$ and the target image $x_t$ extracted by each surrogate are denoted as follows:
\begin{equation} \begin{aligned} \label{eq:get_tokens}
f_{\theta_s}(\tilde{x})
&=[\tilde{\mathbf{z}}_{\mathrm{cls}}^{s}; \tilde{\mathbf{z}}_{1}^{s},\ldots,\tilde{\mathbf{z}}_{N_s}^{s}]
= [\tilde{\mathbf{z}}_{\mathrm{cls}}^{s}; \tilde{\mathbf{Z}}_{\mathrm{patch}}^{s}] \in\mathbb{R}^{(N_s+1)\times d_s}, \\
f_{\theta_s}(x_t)
&= [\mathbf{z}_{\mathrm{cls}}^{t,s};
\mathbf{z}_{1}^{t,s},\ldots,\mathbf{z}_{N_s}^{t,s}]
= [\mathbf{z}_{\mathrm{cls}}^{t,s}; \mathbf{Z}_{\mathrm{patch}}^{t,s}] \in\mathbb{R}^{(N_s+1)\times d_s},
\end{aligned} \end{equation}
where $N_s$ and $d_s$ denote the number of patch tokens and the token embedding dimension associated with the $s$-th surrogate encoder, respectively.
Accordingly,  the collaborative importance matrix over its token sequence can be calculated as:
\begin{equation}\label{eq:get_importance_m}
\tilde{\mathbf{R}}^{s}
\in \mathbb{R}^{(N_s + 1) \times (N_s + 1)},
\end{equation}
where the importance scores associated with the class token are denoted as:
\begin{equation}
\tilde{r}^s=\tilde{\mathbf{R}}^s_{\mathrm{cls}, 1: N_s}\in \mathbb{R}^{N_s}.
\end{equation}
Then, each patch-level importance vector is reshaped into a two-dimensional importance map with the same spatial layout as the patch grid of $f_{\theta_s}(\cdot)$:
\begin{equation}
\tilde{\mathbf{U}}^s
=\mathrm{Reshape}
(\tilde{r}^s;H_s,W_s)
\in \mathbb{R}^{H_s\times W_s},
\end{equation}
where $H_sW_s=N_s$.
To mitigate the spatial granularity discrepancies induced by different input patch sizes across surrogate encoders, the ensemble importance maps are subsequently interpolated to the input-image resolution and aggregated as:
\begin{equation} \label{eq:get_aggregated_m}
\mathbf{U}
=\frac{1}{S} \sum_{s=1}^{S}
\mathrm{Interp} \left(\tilde{\mathbf{U}}^{s}; H, W \right) \in \mathbb{R}^{H \times W}.
\end{equation}
where $\mathrm{Interp}(\cdot; H, W )$ denotes bilinear interpolation that resizes each importance map to the height $H$ and width $W$ of the input image.
Afterward, when identifying critical adversarial patch tokens from the patch-token sequence of each surrogate encoder, the aggregated importance map is resampled to the spatial resolution of the corresponding patch grid and flattened into a surrogate-specific importance vector:
\begin{equation} \label{eq:flatten}
\mathbf{r}_{\mathrm{agg}}^{s}
=\mathrm{Flatten} \left(\mathrm{Interp}\left( \mathbf{U}; H_s, W_s\right) \right) \in \mathbb{R}^{N_s},
\end{equation}
and the surrogate-specific perceptual representations can be obtained following Eqs.~(\ref{select_patch}) and (\ref{get_patch}):
\begin{equation} \label{eq:get_per}
\tilde{\mathbf{Z}}_{\mathrm{per}}^{s}
= \tilde{\mathbf{Z}}_{\mathrm{patch}}^{s} \left[ \mathcal{I}_{\rho}^{s}\right],
\quad
\mathcal{I}_{\rho}^{s}=\mathrm{Top}_{\rho}
\left(\mathbf{r}_{\mathrm{agg}}^{s} \right).
\end{equation}
Subsequently, cross-attentive local perceptual adaptation over the surrogate ensemble is performed through:
\begin{equation} \label{eq:get_per_loss}
\mathcal{L}_{\mathrm{per}}^{\mathrm{ens}}
=1-\frac{1}{S}\sum_{s=1}^{S}\frac{1}{N_{\rho}^{s}} \sum_{i=1}^{N_{\rho}^{s}}
\frac{\left\langle \tilde{\mathbf{z}}_{\mathrm{per}, i}^{s}, \mathbf{z}_{\mathrm{ref}, i}^{s} \right\rangle}{
\left\|\tilde{\mathbf{z}}_{\mathrm{per}, i}^{s} \right\|_2 \left\| \mathbf{z}_{\mathrm{ref}, i}^{s}\right\|_2},
\end{equation}
where $N_{\rho}^s={\lceil \rho N_s \rceil}$, and $\mathbf{z}_{\mathrm{ref}, i}^{s}$ denotes the surrogate-specific target semantic reference associated with each selected adversarial patch token.
In parallel, ensemble-level global conceptual calibration is carried out by:
\begin{equation} \label{eq:get_con_loss}
\mathcal{L}_{\mathrm{con}}^{\mathrm{ens}}
=1-\frac{1}{S}\sum_{s=1}^{S}
\frac{\left\langle\tilde{\mathbf{z}}_{\mathrm{cls}}^{s},\mathbf{z}_{\mathrm{cls}}^{t,s}\right\rangle}
{\left\|\tilde{\mathbf{z}}_{\mathrm{cls}}^{s} \right\|_2 \left\|\mathbf{z}_{\mathrm{cls}}^{t,s} \right\|_2}.
\end{equation}
Finally, the joint optimization for adversarial example generation can be represented as a weighted objective:
\begin{equation} \label{eq:get_joint_loss}
\mathcal{L}_{\mathrm{joint}} =\mathcal{L}_{\mathrm{con}}^{\mathrm{ens}}
+ \lambda \mathcal{L}_{\mathrm{per}}^{\mathrm{ens}},
\end{equation}
where $\lambda$ is used to control the loss contributions of conceptual calibration and perceptual adaptation.

\begin{algorithm}[t] 
\label{alg} 
\caption{The GeoThreat Attack Algorithm}
\SetKwInput{KwInput}{Input}   
\SetKwInput{KwOutput}{Output}    
\DontPrintSemicolon
\KwInput{Clean image $x$ and target image $x_t$,  surrogate model ensemble $\mathcal{F}=\{f_{\theta_s}\}_{s=1}^{S}$,
patch selection ratio $\rho$, temperature parameter $\tau$, weighting coefficient $\lambda$, perturbation magnitude $\epsilon$, number of iterations $K$, step size $\alpha$, and momentum coefficient $\mu$.}
\KwOutput{The adversarial example $\tilde{x}$.}
Initialize: $\tilde{x}^{0} = x$, ${g}^{0} = 0$.\;
\For{$k = 1$ \KwTo $K$}{
Apply random block shuffling to $\tilde{x}^{k-1}$;\;
\For {$s = 1$ \KwTo $S$}{
Extract the token-wise visual representations of $\tilde{x}^{k-1}$ and $x_t$ with $f_{\theta}^{s}$ by Eq.~(\ref{eq:get_tokens});\;
Estimate the collaborative importance matrix $\tilde{\mathbf{R}}^{s}$ by Eq.~(\ref{eq:get_importance_m});\;
}
Aggregate the importance maps by Eq.~(\ref{eq:get_aggregated_m});\;
\For{$s = 1$ \KwTo $S$}{
Get the surrogate-specific perceptual representations by Eqs.~(\ref{eq:flatten}) and (\ref{eq:get_per});\;
Estimate the target semantic references by Eqs.~(\ref{eq:get_cross_a}) - (\ref{eq:get_ref});\;
}
Compute $\mathcal{L}_{\mathrm{joint}}$ by Eqs.~(\ref{eq:get_per_loss}) - (\ref{eq:get_joint_loss});\;
$g^k = \mu \cdot g^{k-1} + \nabla_{\tilde{x}^{k-1}}\mathcal{L}_{joint}$;\;
$\tilde{x}^{k}=\operatorname{Clip}_{x, \epsilon}(\tilde{x}^{k-1}-\alpha \cdot \operatorname{sign}(g^k))$;\;
}
\textbf{end for}\\
\Return{$\tilde{x}=\tilde{x}^{K}$}\;
\end{algorithm}

\section{Experiments}\label{sec4}
\subsection{Experimental Setup}
\subsubsection{Dataset}
We conduct experiments on three widely used remote sensing image datasets, UCM~\cite{yang2010ucm}, SIRI-WHU~\cite{zhao2015siri}, and AID~\cite{xia2017aid}, to evaluate the targeted attack performance of different methods against LVLMs for remote sensing image interpretation.
Specifically, the evaluation involves two major remote sensing image interpretation tasks, i.e., image captioning and image classification.
For the image captioning task, we randomly construct 500 image pairs from the UCM and SIRI-WHU datasets, with each pair comprising images from different land-cover types that are alternately treated as the source and target images during adversarial attack generation.
For the image classification task, we select ten fixed source-target class pairs from the AID dataset, with 50 images sampled from each class, and aim to mislead LVLM predictions on source images toward the corresponding target classes under a zero-shot setting.
During the experiments, the input resolution of all images is set to $224 \times 224$. 

\subsubsection{Model Architectures}
In our experiments, we adopt five CLIP variants~\cite{radford2021learning} as ensemble surrogate models for adversarial example generation, including ViT-B/32, ViT-B/16, ViT-L/14, ViT-H/14, and ViT-g/14.
We evaluate the transferable targeted attack performance across seven LVLMs, covering three general-purpose open-source models (InstructBLIP~\cite{dai2023instructblip}, LLaVA-1.5-7B~\cite{liu2023visual}, and Qwen2.5-VL-7B~\cite{bai2025qwen25vl}), two commercial proprietary models (GPT-5-mini~\cite{openai2025gpt5} and Gemini-2.5-Flash~\cite{comanici2025gemini25}, and two remote sensing-specific models (GeoChat~\cite{kuckreja2024geochat} and TeoChat~\cite{irvin2025teochat}).
For the image captioning and classification tasks, the text prompts for these models are set to “Describe the image in one concise sentence.” and “This image belongs to which remote sensing scene category in the following category list?”, respectively.

\subsubsection{General Implementation}
During iterative perturbation generation, all attack methods are implemented under a unified setting to ensure a fair comparison.
Specifically, the maximum perturbation magnitude $\epsilon$ under the $\ell_{\infty}$-norm constraint is set to $16/255$, and adversarial examples are optimized for 300 iterations with a step size of $\alpha=1/255$.
For the hyperparameter configuration of GeoThreat, the selection ratio $\rho$ of adversarial patch tokens, temperature parameter $\tau$, and loss weighting coefficient $\lambda$ are set to $0.5$, $0.1$, and $0.5$, respectively.
The experiments are implemented within the PyTorch framework and executed on NVIDIA RTX 4090 GPUs.

\subsubsection{Evaluation Metrics}
To evaluate the effectiveness of the proposed method, we adopt the following metrics to assess the targeted attack performance against LVLMs for remote sensing image interpretation across different tasks.
For attacks on the image captioning task, we follow the commonly used LLM-as-a-judge protocol~\cite{li2025mattack,jia2025foa}.
Specifically, we measure the semantic similarity between the textual descriptions generated by the same model for the adversarial example and the corresponding target image using GPTScore~\cite{fu2024gptscore}.
The attack is considered successful when the similarity score exceeds 0.5. 
The attack performance is quantified by the overall attack success rate~(ASR) and the average similarity score~(AvgSim).
For the image classification task, attack success is instead defined at the label level, i.e., whether the prediction of the adversarial example can be shifted from its ground-truth class toward the predefined target class.

\subsection{Experimental Results}
To demonstrate the effectiveness of our proposed method, we compare the attack performance of GeoThreat with eight advanced adversarial attack methods for LVLMs, including AttackVLM~\cite{zhao2023mf-ii}, AdvDiffVLM~\cite{guo2024efficient}, AnyAttack~\cite{zhang2025anyattack}, SSA-CWA~\cite{dong2023robust}, M-Attack~\cite{li2025mattack}, VEAttack~\cite{mei2026veattack}, FOA-Attack~\cite{jia2025foa}, and V-Attack~\cite{nie2026vattack}.

\begin{table*}[htbp]
\centering
\small
\setlength{\tabcolsep}{4pt}
\renewcommand{\arraystretch}{1}
\caption{Transferable targeted attack performance on image captioning under the UCM $\rightarrow$ SIRI-WHU setting.}
\begin{tabular}{l|c|cc|cc|cc|cc|cc|cc}
\toprule
\multirow[c]{2}{*}[-0.8ex]{\textbf{Method}} & 
\multirow[c]{2}{*}[-0.8ex]{\textbf{Model}} & 
\multicolumn{2}{c|}{\textbf{InstructBlip}} & 
\multicolumn{2}{c|}{\textbf{LLaVa-1.5-7B}} & 
\multicolumn{2}{c|}{\textbf{Qwen2.5-VL-7B}} & 
\multicolumn{2}{c|}{\textbf{GPT-5-mini}} & 
\multicolumn{2}{c|}{\textbf{GeoChat}} & 
\multicolumn{2}{c}{\textbf{TeoChat}} \\
\cmidrule(lr){3-4} \cmidrule(lr){5-6} \cmidrule(lr){7-8} \cmidrule(lr){9-10} \cmidrule(lr){11-12} \cmidrule(lr){13-14}
& &\textbf{ASR} & \textbf{AvgSim} &\textbf{ASR} & \textbf{AvgSim} &\textbf{ASR} & \textbf{AvgSim} & \textbf{ASR} & \textbf{AvgSim} & \textbf{ASR} & \textbf{AvgSim} & \textbf{ASR} & \textbf{AvgSim} \\ \midrule
Clean
& / & 0.0  & 0.16 & 0.0  & 0.08 & 0.0  & 0.15 & 0.0  & 0.16 & 0.0  & 0.13 & 0.0  & 0.14  \\ \midrule
\multirow[c]{6}{*}{AttackVLM~\cite{zhao2023mf-ii}}
& B/32  & 5.8  & 0.19  & 2.4  & 0.11  & 13.4 & 0.23 & 11.0 & 0.21  & 5.0  & 0.20  & 7.0  & 0.19\\
& B/16  & 8.0  & 0.21  & 2.4  & 0.12  & 13.0 & 0.24 & 11.8 & 0.22  & 3.8  & 0.20  & 5.4  & 0.18\\
& L/14 & 8.0  & 0.20 & 4.6  & 0.14 & 10.6 & 0.22 & 2.4  & 0.18 & 11.0 & 0.22 & 8.6  & 0.20\\
& H/14 & 22.0 & 0.31 & 6.2  & 0.15 & 25.2 & 0.32 & 4.6  & 0.20 & 16.0 & 0.24 & 23.0 & 0.29\\
& g/14
& 20.8 & 0.30
& 4.4  & 0.13
& 24.6 & 0.33
& 4.0  & 0.20
& 13.0 & 0.22
& 21.6 & 0.28\\
& Ensemble
& 42.0 & 0.44
& 18.6 & 0.25
& 36.0 & 0.40
& 8.8  & 0.24
& 30.2 & 0.34
& 38.2 & 0.41\\
\midrule
AdvDiffVLM~\cite{guo2024efficient}
& Ensemble
& 3.0  & 0.16
& 1.4  & 0.10
& 4.6  & 0.14
& 2.0  & 0.17
& 2.8  & 0.17
& 4.6  & 0.17\\
AnyAttack~\cite{zhang2025anyattack}
& Ensemble
& 2.8  & 0.17
& 1.0  & 0.10
& 5.2  & 0.18
& 2.0  & 0.17
& 4.0  & 0.16
& 4.2  & 0.17\\
SSA-CWA~\cite{dong2023robust}
& Ensemble
& 64.4 & 0.59
& 47.2 & 0.48
& 64.8 & 0.60
& 29.6 & 0.36
& 45.2 & 0.44
& 59.8 & 0.55\\
M-Attack~\cite{li2025mattack}
& Ensemble
& 46.6 & 0.46
& 21.8 & 0.27
& 40.4 & 0.43
& 11.8 & 0.26
& 32.8 & 0.36
& 42.0 & 0.43\\
VEAttack~\cite{mei2026veattack}
& Ensemble
& 40.6 & 0.42
& 14.4 & 0.22
& 32.8 & 0.38
& 9.4  & 0.23
& 26.6 & 0.32
& 44.6 & 0.43\\
FOA-Attack~\cite{jia2025foa}
& Ensemble
& 57.4 & 0.52
& 23.8 & 0.30
& 48.6 & 0.47
& 15.2 & 0.27
& 37.4 & 0.40
& 53.4 & 0.50\\
V-Attack~\cite{nie2026vattack}
& Ensemble
& 23.0 & 0.28
& 10.2 & 0.21
& 19.8 & 0.25
& 13.2 & 0.22
& 22.4 & 0.28
& 15.8 & 0.25\\
\midrule
\rowcolor{red!8} \textbf{GeoThreat}
& Ensemble
& \textbf{88.0} & \textbf{0.73}
& \textbf{73.2} & \textbf{0.62}
& \textbf{78.6} & \textbf{0.67}
& \textbf{50.0} & \textbf{0.50}
& \textbf{72.0} & \textbf{0.63}
& \textbf{70.8} & \textbf{0.61}\\
\bottomrule
\end{tabular}
\label{tab:main_results}
\end{table*}

\begin{table*}[htbp]
\centering
\small
\setlength{\tabcolsep}{4pt}
\renewcommand{\arraystretch}{1}
\caption{Transferable targeted attack performance on image captioning under the SIRI-WHU $\rightarrow$ UCM setting. }

\begin{tabular}{l|c|cc|cc|cc|cc|cc|cc}
\toprule
\multirow[c]{2}{*}[-0.8ex]{\textbf{Method}} &
\multirow[c]{2}{*}[-0.8ex]{\textbf{Model}} &
\multicolumn{2}{c|}{\textbf{InstructBlip}} &
\multicolumn{2}{c|}{\textbf{LLaVa-1.5-7B}} &
\multicolumn{2}{c|}{\textbf{Qwen2.5-VL-7B}} &
\multicolumn{2}{c|}{\textbf{GPT-5-mini}} &
\multicolumn{2}{c|}{\textbf{GeoChat}} &
\multicolumn{2}{c}{\textbf{TeoChat}}\\
\cmidrule(lr){3-4}
\cmidrule(lr){5-6}
\cmidrule(lr){7-8}
\cmidrule(lr){9-10}
\cmidrule(lr){11-12}
\cmidrule(lr){13-14}
& &
\textbf{ASR} & \textbf{AvgSim} &
\textbf{ASR} & \textbf{AvgSim} &
\textbf{ASR} & \textbf{AvgSim} &
\textbf{ASR} & \textbf{AvgSim} &
\textbf{ASR} & \textbf{AvgSim} &
\textbf{ASR} & \textbf{AvgSim} \\
\midrule

Clean
& /
& 0.0  & 0.16
& 0.0  & 0.09
& 0.0  & 0.16
& 0.0  & 0.17
& 0.0  & 0.14
& 0.0  & 0.15\\
\midrule
\multirow[c]{6}{*}{AttackVLM~\cite{zhao2023mf-ii}}
& B/32
& 8.2  & 0.22
& 2.2  & 0.13
& 12.8 & 0.24
& 6.6  & 0.22
& 7.6  & 0.19
& 6.6  & 0.19\\
& B/16
& 7.2  & 0.21
& 2.8  & 0.14
& 11.2 & 0.24
& 4.4  & 0.21
& 8.8  & 0.20
& 7.2  & 0.19 \\
& L/14
& 21.8 & 0.30
& 12.2 & 0.23
& 19.0 & 0.30
& 5.8  & 0.21
& 20.8 & 0.28
& 13.8 & 0.26\\
& H/14
& 36.6 & 0.40
& 10.6 & 0.21
& 33.6 & 0.39
& 6.4  & 0.22
& 15.4 & 0.25
& 36.2 & 0.40\\
& g/14
& 33.6 & 0.38
& 7.2  & 0.18
& 28.2 & 0.36
& 6.2  & 0.22
& 13.6 & 0.24
& 32.6 & 0.38\\
& Ensemble
& 64.8 & 0.57
& 36.4 & 0.38
& 51.6 & 0.51
& 18.2 & 0.31
& 39.8 & 0.39
& 54.6 & 0.52\\
\midrule

AdvDiffVLM~\cite{guo2024efficient}
& Ensemble
& 1.8  & 0.16
& 1.0  & 0.10
& 3.6  & 0.17
& 1.8  & 0.17
& 2.6  & 0.13
& 3.6  & 0.16\\
AnyAttack~\cite{zhang2025anyattack}
& Ensemble
& 3.2   & 0.17
& 1.0  & 0.10
& 5.2  & 0.18
& 1.6  & 0.17
& 4.0  & 0.16
& 4.2  & 0.17\\
SSA-CWA~\cite{dong2023robust}
& Ensemble
& 72.6 & 0.67
& 54.8 & 0.54
& 70.2 & 0.63
& 39.4 & 0.42
& 55.2 & 0.53
& 60.4 & 0.57 \\
M-Attack~\cite{li2025mattack}
& Ensemble
& 60.8 & 0.55
& 36.4 & 0.39
& 52.8 & 0.52
& 20.4 & 0.32
& 38.6 & 0.41
& 56.0 & 0.52 \\
VEAttack~\cite{mei2026veattack}
& Ensemble
& 56.4 & 0.52
& 24.8 & 0.31
& 42.8 & 0.45
& 15.8 & 0.29
& 30.6 & 0.36
& 53.2 & 0.53 \\
FOA-Attack~\cite{jia2025foa}
& Ensemble
& 69.4 & 0.61
& 44.4 & 0.44
& 59.6 & 0.55
& 24.8 & 0.36
& 48.4 & 0.46
& 61.4 & 0.57 \\
V-Attack~\cite{nie2026vattack}
& Ensemble
& 47.8 & 0.46
& 32.8 & 0.36
& 37.4 & 0.38
& 33.4 & 0.36
& 44.8 & 0.42
& 38.4 & 0.42 \\
\midrule
\rowcolor{red!8} \textbf{GeoThreat}
& Ensemble
& \textbf{88.0} & \textbf{0.76}
& \textbf{80.8} & \textbf{0.69}
& \textbf{83.0} & \textbf{0.72}
& \textbf{70.6} & \textbf{0.61}
& \textbf{75.0} & \textbf{0.64}
& \textbf{72.6} & \textbf{0.64} \\
\bottomrule
\end{tabular}
\label{tab:main_results_2}
\end{table*}

\begin{figure}[]
\centering 
\includegraphics[width=0.5\textwidth]{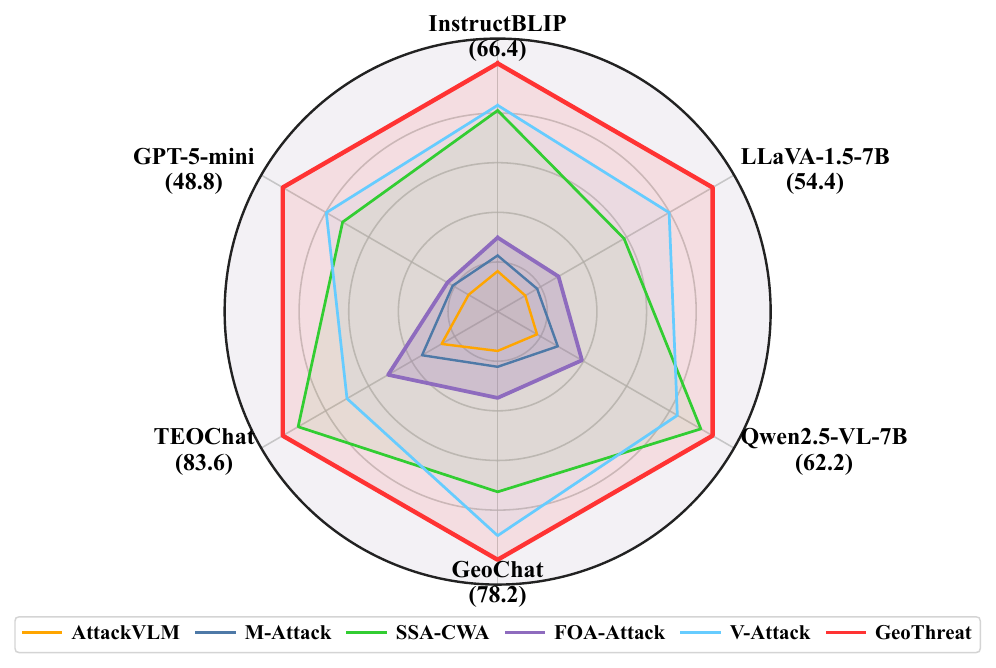}
\caption{Attack success rates of different methods across various LVLMs on the remote sensing image classification task based on the AID dataset.}
\label{classification_res} 
\end{figure}

\begin{figure*}[]
\centering 
\includegraphics[width=\textwidth]{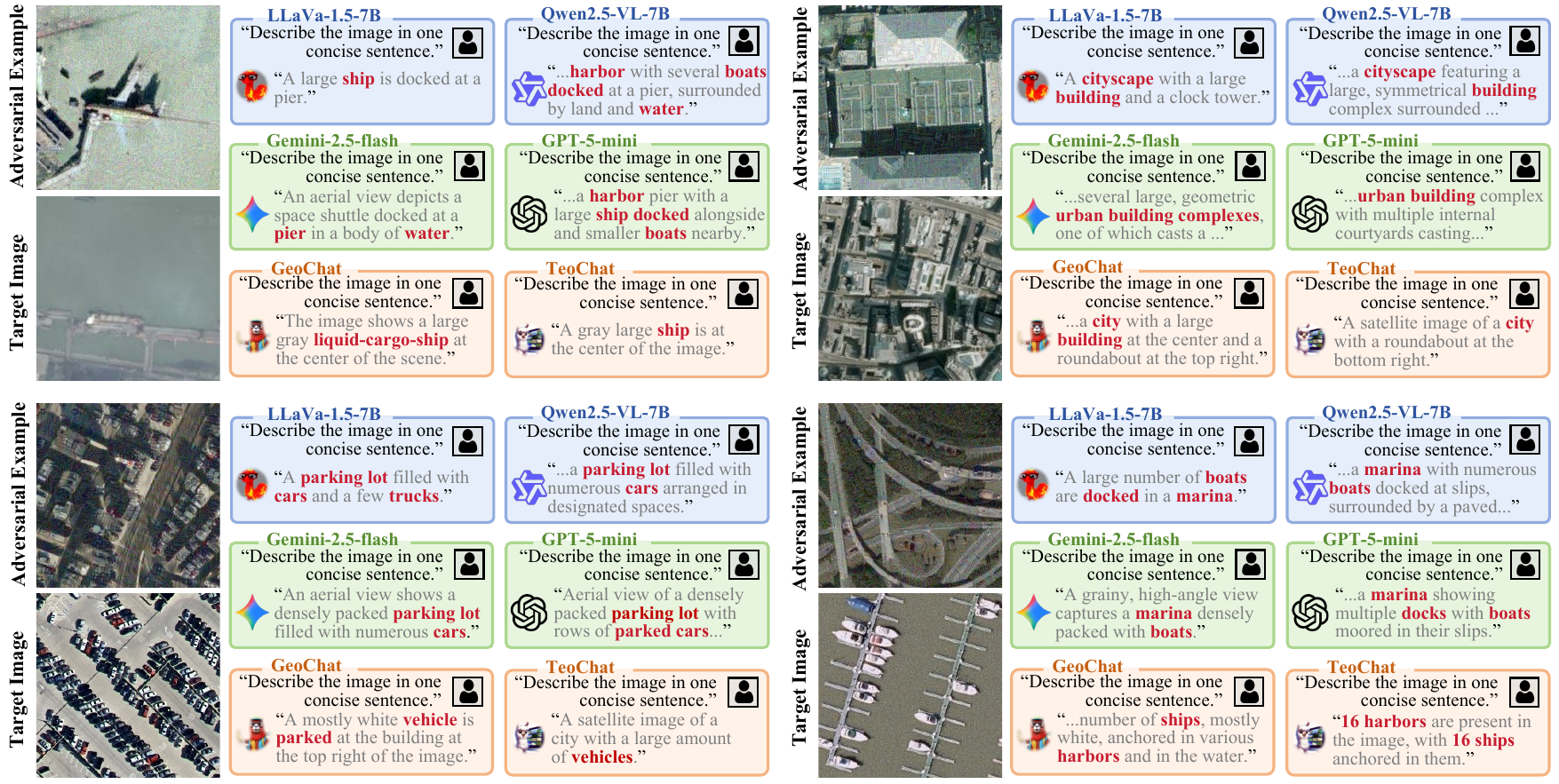}
\caption{Illustration of GeoThreat for transferable targeted adversarial attacks against LVLMs in the remote sensing image captioning task. Target-related semantic errors are highlighted in red.}
\label{caption_vis} 
\end{figure*}

\subsubsection{Quantitative Analysis}
The comparison results of transferable targeted attacks on image captioning under different source-target settings are presented in Tables \ref{tab:main_results} and \ref{tab:main_results_2}.
The first row of each table reports the attack-free results on clean inputs, which serve as a reference for more clearly assessing the extent of target-oriented semantic manipulation induced by each adversarial attack method.
According to the results, GeoThreat consistently achieves the strongest attack performance across all evaluated LVLMs, outperforming existing methods by a substantial margin.
Specifically, on generic open-source LVLMs, GeoThreat achieves an average ASR of 81.9\%, outperforming recent state-of-the-art LVLM attack methods VEAttack and V-Attack by 46.6\% and 53.4\%, respectively.
Such superiority in transferability and controllability remains evident on remote sensing-specific LVLMs, on which GeoThreat still achieves ASRs exceeding 70\%.
Moreover, compared with FOA-Attack, which also exploits local representations for black-box semantic manipulation, GeoThreat achieves an overall AvgSim above 0.65 across different LVLMs, clearly surpassing the approximately 0.45 obtained by FOA-Attack.
This demonstrates that our proposed method can more effectively adapt local perceptual cues toward the designated semantics.
Notably, among the compared methods, SSA-CWA achieves stronger attack performance on LVLM-based remote sensing image interpretation than several advanced attacks that are more effective in natural image understanding scenarios.
This is mainly attributed to the block-wise random shuffling operation, which encourages the exploration of local perceptual cues across diverse spatial layouts during adversarial example generation.
Furthermore, the use of model ensemble techniques offers clear advantages over a single surrogate model in achieving transferable targeted adversarial attack, as it helps reduce surrogate-specific biases and provides more generalizable guidance for perturbation update.

In addition to the image captioning task, we further report the targeted attack performance of different methods against LVLMs on the image classification task, with the results presented in Fig.~\ref{classification_res}.
We can see that GeoThreat uniformly achieves the highest ASRs across different LVLMs, indicating its strong capability to mislead model predictions toward the predefined target classes.
This further confirms its superiority in both transferability and controllability when inducing targeted semantic manipulation.
For the other methods, V-Attack exhibits relatively balanced and competitive attack performance on the image classification task.
This can be mainly attributed to the accurate identification of concept-relevant local value features, which facilitates target-oriented mapping of category semantics.
Regarding the victim LVLMs, the remote sensing-specific models, i.e., GeoChat and TeoChat, exhibit more pronounced vulnerability to adversarial perturbations.
This further reveals potential reliability risks in LVLM-based remote sensing image interpretation, underscoring the necessity of investigating the adversarial robustness of LVLMs in the remote sensing domain.

\begin{figure*}[]
\centering 
\includegraphics[width=\textwidth]{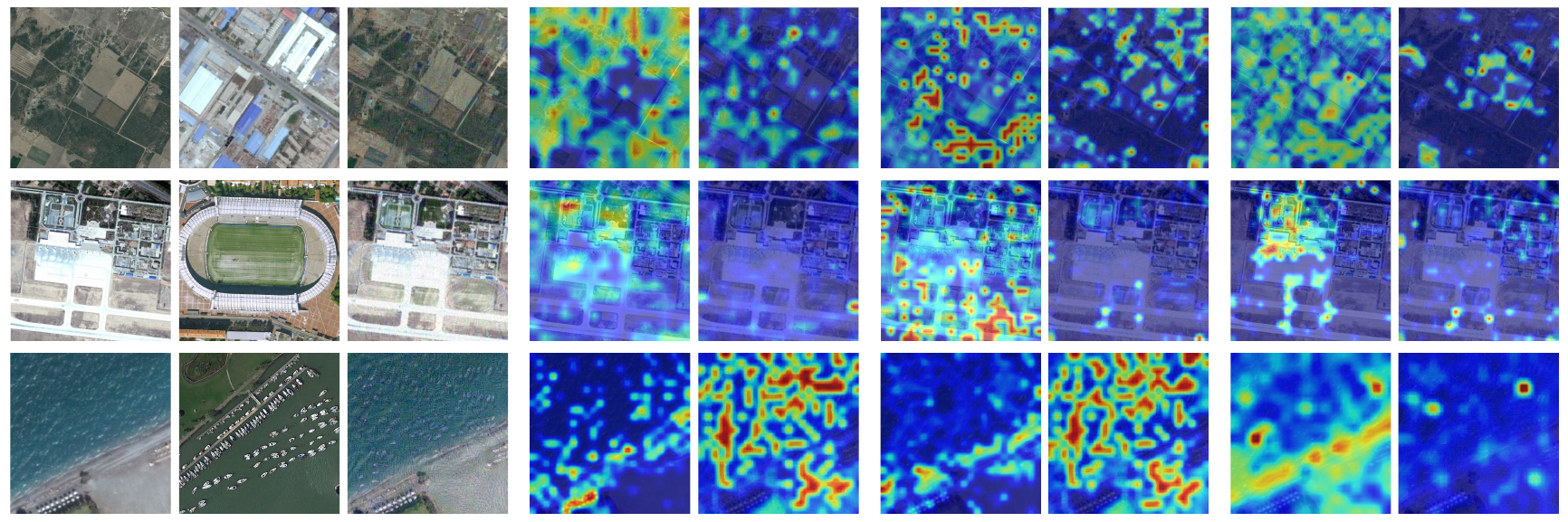}
\caption{Visualization of the attention shift induced by the proposed GeoThreat on the image classification task. The first three columns correspond to the clean input image, target image, and adversarial example, respectively, while the remaining paired columns show the Grad-CAM responses of InstructBLIP, LLaVA-1.5-7B, and GeoChat on clean inputs and adversarial examples.}
\label{cls_vis} 
\end{figure*}

\subsubsection{Qualitative Analysis}
Beyond the quantitative analysis above, we further conduct qualitative analysis to more intuitively demonstrate how GeoThreat misleads LVLM-generated interpretation results for remote sensing images toward specified responses under black-box settings.

First, we present adversarial examples generated by GeoThreat, together with the corresponding captioning results from different LVLMs, in Fig.~\ref{caption_vis}.
The outputs of open-source, commercial closed-source, and remote sensing-specific LVLMs are displayed in blue, green, and orange panels, respectively.
According to the results, by introducing visually imperceptible perturbations, GeoThreat can successfully steer the interpretation results of different LVLMs toward the corresponding semantics of the target images.
For example, given an input image of a residential area, both the overall scene description and the key object identification are redirected to parking lots and cars.
This indicates the effectiveness of the proposed attack method in shifting the joint local-global reasoning process for remote sensing image interpretation.

Then, we further conduct a Grad-CAM-based qualitative analysis to examine the transferable targeted attack behavior of GeoThreat on the remote sensing image classification task.
Specifically, in Fig.~\ref{cls_vis}, we visualize clean input images, corresponding target images, and adversarial examples generated by GeoThreat, together with the Grad-CAM of InstructBLIP, LLaVA-1.5-7B, and GeoChat during scene-category prediction.
Although the visual differences between clean and adversarial images are subtle, the attention of different LVLMs is largely shifted away from the original evidence supporting correct classification toward irrelevant regions.
As a representative case illustrated in the last row, different victim LVLMs attend to water-surface regions that serve as primary evidence for port recognition in the adversarial example, instead of the shoreline regions that support beach recognition under clean conditions.

\subsection{Ablation Studies}
\subsubsection{Contribution of Each Part}
GeoThreat involves three main parts during adversarial example generation: joint modulation of conceptual and perceptual representations, collaborative importance estimation~(CIE), and cross-attentive perceptual adaptation~(CPA).
In this subsection, we conduct ablation experiments on the image captioning task to validate the contribution of each part to enabling transferable and controllable semantic manipulation.
Specifically, under the UCM $\rightarrow$ SIRI-WHU setting, we evaluate the attack performance of GeoThreat under different configurations of the relevant parts, using LLaVA-1.5-7B, TeoChat, and GPT-5-mini as victim models.

\begin{table}[]
\centering
\small 
\setlength{\tabcolsep}{3.5pt}
\renewcommand{\arraystretch}{1}
\caption{Ablation study on joint conceptual-perceptual representation modulation.}
\label{ab_loss}
\begin{tabular}{l|cc|cc|cc}
\toprule
\multirow{2}{*}[-0.8ex]{\textbf{Strategy}} &  \multicolumn{2}{c|}{\textbf{LLaVa-1.5-7B}} & \multicolumn{2}{c|}{\textbf{TeoChat}} & \multicolumn{2}{c}{\textbf{GPT-5-mini}} \\
\cmidrule(lr){2-3} \cmidrule(lr){4-5} \cmidrule(lr){6-7} & \textbf{ASR} & \textbf{AvgSim} & \textbf{ASR} & \textbf{AvgSim} & \textbf{ASR} & \textbf{AvgSim} \\
\midrule
\textbf{Conceptual} & 61.8 & 0.57  & 66.6 & 0.58  & 42.2 & 0.45   \\
\textbf{Perceptual} & 53.8 & 0.51 & 65.2 & 0.57 & 36.6 & 0.41 \\
\textbf{Joint} & \textbf{73.2} & \textbf{0.62} & \textbf{70.8} & \textbf{0.61} & \textbf{50.0} & \textbf{0.50} \\
\bottomrule
\end{tabular}
\end{table}

\begin{table}[b]
\centering \small 
\setlength{\tabcolsep}{4pt}
\renewcommand{\arraystretch}{1}
\caption{Ablation experiments about contributions of CIE and CPA.}
\begin{tabular} {cc|cc|cc|cc}
\toprule
\multirow[c]{2}{*}[-0.8ex]{\textbf{CIE}} &  \multirow[c]{2}{*}[-0.8ex]{\textbf{CPA}} & \multicolumn{2}{c|}{\textbf{LLaVa-1.5-7B}} & \multicolumn{2}{c|}{\textbf{TeoChat}} & \multicolumn{2}{c} {\textbf{GPT-5-mini}} \\
\cmidrule(lr){3-4} \cmidrule(lr){5-6} \cmidrule(lr){7-8}
& & \textbf{ASR} & \textbf{AvgSim} & \textbf{ASR} & \textbf{AvgSim} & \textbf{ASR} & \textbf{AvgSim} \\
\midrule
\ding{56}& \ding{56} & 64.6 & 0.56   & 67.0 & 0.57  & 42.6 & 0.45 \\
\ding{52}& \ding{56} & 69.2 & 0.60  & 68.2 & 0.58  & 45.0 & 0.46 \\
\ding{56}& \ding{52} & 66.8 & 0.59  & 66.6 & 0.57  & 44.2 & 0.48  \\
\ding{52}& \ding{52} & \textbf{73.2} & \textbf{0.62} & \textbf{70.8} & \textbf{0.61} & \textbf{50.0} & \textbf{0.50} \\
\bottomrule
\end{tabular}
\label{tab_percept}
\end{table}

First, we validate the effectiveness of joint conceptual-perceptual representation modulation, which serves as our central design rationale for tackling the challenge arising from the distinctive reasoning paradigm of remote sensing image interpretation in achieving transferable targeted attacks.
The attack performance against different LVLMs is reported in Table~\ref{ab_loss}, where using either global conceptual matching or local perceptual alignment alone to guide adversarial example generation leads to a clear performance gap compared with their joint use.
Therefore, the adoption of joint representation modulation is essential for enhancing targeted black-box attack efficacy against LVLM-based remote sensing interpretation.
Then, we compare the attack performance of different combinations of the remaining two parts related to local perceptual adaptation, with the results reported in Table~\ref{tab_percept}.
We adopt a uniform alignment scheme between all patch tokens of the adversarial example and those of the target image as the baseline.
We observe that the individual use of each part brings only moderate improvements, whereas their coordination yields substantially stronger transferable targeted attack effects.
This indicates that accurate identification of critical adversarial patches and appropriate modeling of target correspondences are both indispensable for adapting local perceptual cues toward the designated semantics.

\begin{table}[]
\centering \small
\setlength{\tabcolsep}{3.5pt}
\renewcommand{\arraystretch}{1}
\caption{Ablation experiments about the identification approach of critical adversarial patch tokens.}
\label{tab_select} \begin{tabular}{l|cc|cc|cc}
\toprule
\multirow{2}{*}[-0.8ex]{\textbf{Approach}} &  \multicolumn{2}{c|}{\textbf{LLaVa-1.5-7B}} & \multicolumn{2}{c|}{\textbf{TeoChat}} & \multicolumn{2}{c}{\textbf{GPT-5-mini}} \\
\cmidrule(lr){2-3} \cmidrule(lr){4-5} \cmidrule(lr){6-7} & \textbf{ASR} & \textbf{AvgSim} & \textbf{ASR} & \textbf{AvgSim} & \textbf{ASR} & \textbf{AvgSim} \\
\midrule
\textbf{Random} & 66.6 & 0.57  & 67.2 & 0.58  & 45.4 & 0.47   \\
\textbf{Attention} & 67.2 & 0.57 & 68.4 & 0.59 & 45.6 & 0.46 \\
\textbf{Gradient} & 67.8 & 0.58 & 69.8 & 0.60 & 49.6 & 0.49 \\
\textbf{Collabrative} & \textbf{73.2} & \textbf{0.62} & \textbf{70.8} & \textbf{0.61} & \textbf{50.0} & \textbf{0.50} \\
\bottomrule
\end{tabular}
\end{table}

\begin{figure*}[] \centering
\subfloat[$\lambda$]{\includegraphics[width=0.45\textwidth]
{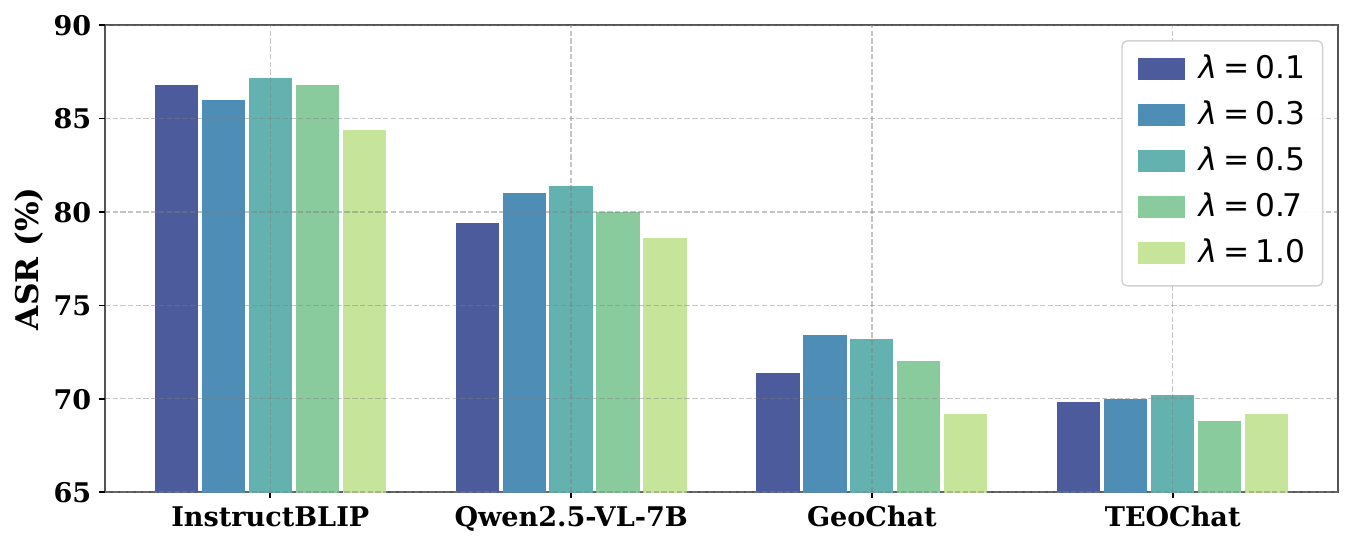}\label{para1}}
\hspace{0.5cm}
\subfloat[$\rho$]{\includegraphics[width=0.45\textwidth]{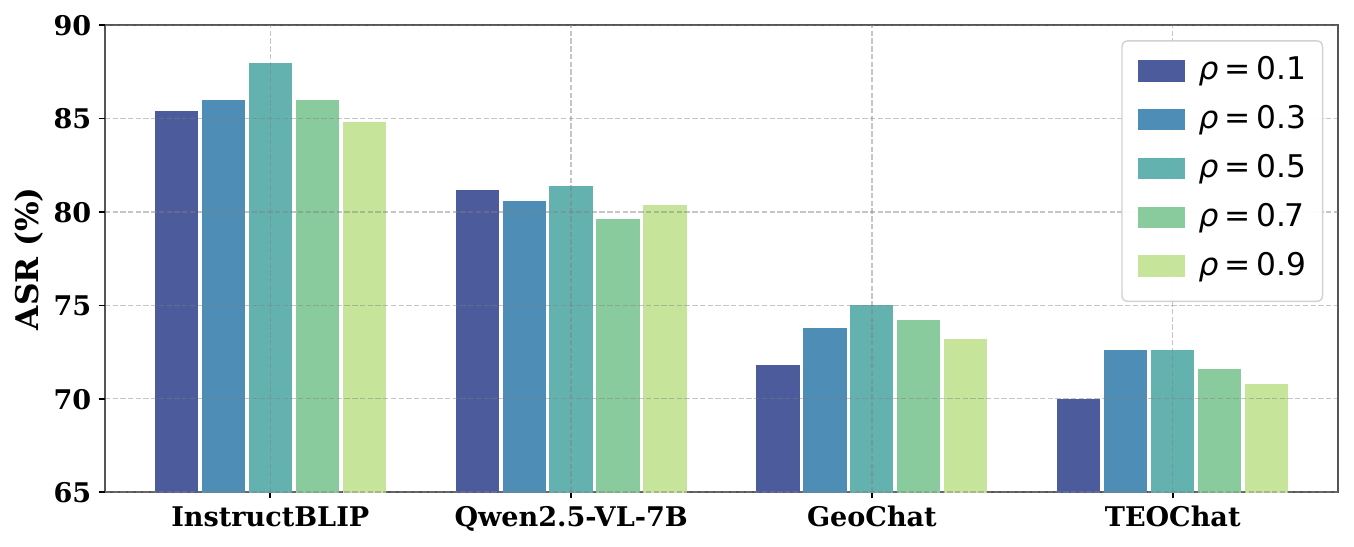}\label{para2}} 
\caption{Attack success rate of adversarial examples generated by GeoThreat with different settings of $\lambda$ and $\rho$.}
\label{parameter_analyze} 
\end{figure*}

\subsubsection{Identification of Critical Adversarial Patch Tokens}
When determining local perceptual representations, we develop a collaborative importance estimation approach to identify adversarial patch tokens that are both decision-relevant and target-responsive.
To justify this design choice in critical adversarial patch identification, we compare the attack performance of GeoThreat under the UCM $\rightarrow$ SIRI-WHU attack protocol, where critical adversarial patches are identified using different importance-estimation approaches.

Table~\ref{tab_select} reports ASR and AvgSim under different importance-estimation approaches. 
Here, Random denotes randomly selecting a fixed proportion of adversarial patch tokens as perceptual representations.
Attention and Gradient denote using rolled-out attention scores and conceptual-similarity sensitivities as the importance-estimation criterion, respectively.
According to the results, selecting adversarial patches solely based on attention scores does not bring evident gains over random selection in driving the target-oriented semantic transition.
In comparison, using patch tokens selected according to conceptual-similarity sensitivities as perceptual representations yields relatively better attack performance.
By integrating attention weights with their corresponding conceptual-similarity sensitivities, the proposed collaborative importance estimation approach can most effectively steer the outputs of different victim models toward the target semantics.

\subsubsection{Hyperparameter Analysis}
In this subsection, we investigate how different settings of key hyperparameters in GeoThreat affect transferable targeted attack performance, including the adversarial patch-token selection ratio $\rho$ and the loss weighting coefficient $\lambda$.
Specifically, under the SIRI-WHU $\rightarrow$ UCM setting, we compare the attack success rates of GeoThreat across different hyperparameter values.

First, to avoid the influence of patch selection, we use all adversarial patch tokens as perceptual representations and conduct joint adversarial optimization under different values of $\lambda$ ranging from 0.1 to 1.0.
The results presented in Fig.~\ref{para1} indicate that a moderate $\lambda$ yields better attack performance, whereas assigning an overly large weight to local perceptual adaptation can compromise global conceptual calibration and lead to performance degradation.
Then, we first fix $\lambda$ to 0.5 and vary $\rho$ from 0.1 to 0.9. 
As shown in Fig.~\ref{para2}, the attack performance against different LVLMs initially increases as $\rho$ becomes larger, but tends to decline when excessive adversarial patch tokens are selected.
This is because the land-cover semantics encoded in the image cannot be sufficiently characterized by a limited number of adversarial patch tokens, whereas excessive patch selection can introduce redundant information and dilute the focus on discriminative cues.
Given the above analysis, we set both $\lambda$ and $\rho$ to 0.5 in GeoThreat.

\section{Conclusion}\label{sec5}
In this paper, we have proposed GeoThreat, a transferable targeted attack for LVLM-based remote sensing image interpretation.
In response to the challenges arising from the joint global-local reasoning paradigm for black-box targeted attacks, GeoThreat guides adversarial example generation through coordinated modulation of conceptual and perceptual representations.
Specifically, the class token of the adversarial example is aligned with that of the target image to perform global conceptual calibration.
Meanwhile, at the perceptual level, critical adversarial patch tokens are identified through collaborative importance estimation and cross-attentively adapted toward the designated semantics.
Finally, adversarial perturbations are iteratively updated through a joint adversarial optimization strategy over an ensemble of surrogate vision encoders to enable more transferable and controllable semantic manipulation.
Extensive experiments on image captioning and image classification across multiple remote sensing datasets demonstrate that GeoThreat consistently outperforms existing attack methods against diverse LVLMs.
Beyond the performance superiority, the established benchmark reveals potential security and reliability risks in LVLM-based remote sensing image interpretation, highlighting the importance of exploring adversarial robustness in the remote sensing field amid the rise of LVLMs.

\bibliographystyle{IEEEtran}
\bibliography{Refs}

@article{liu2023visual,
  title={Visual instruction tuning},
  author={Liu, Haotian and Li, Chunyuan and Wu, Qingyang and Lee, Yong Jae},
  journal={Advances in Neural Information Processing Systems},
  volume={36},
  pages={34892--34916},
  year={2023}
}

@article{sun2024task,
  title={Task-specific importance-awareness matters: On targeted attacks against object detection},
  author={Sun, Xuxiang and Cheng, Gong and Li, Hongda and Peng, Hongyu and Han, Junwei},
  journal={IEEE Transactions on Circuits and Systems for Video Technology},
  volume={34},
  number={11},
  pages={11619--11629},
  year={2024},
  publisher={IEEE}
}

@article{li2025sparse,
  title={Sparse unmixing guided adversarial attack for hyperspectral image classification},
  author={Li, Hao and Dang, Kelin and Gong, Maoguo and Qin, AK and Zhou, Yu and Wu, Yue and Xing, Lining},
  journal={IEEE Transactions on Circuits and Systems for Video Technology},
  year={2025},
  publisher={IEEE}
}

@article{yuan2019adversarial,
  title={Adversarial examples: Attacks and defenses for deep learning},
  author={Yuan, Xiaoyong and He, Pan and Zhu, Qile and Li, Xiaolin},
  journal={IEEE Transactions on Neural Networks and Learning Systems},
  volume={30},
  number={9},
  pages={2805--2824},
  year={2019},
  publisher={IEEE}
}

@article{qi2024masked,
  title={Masked spatial--spectral autoencoders are excellent hyperspectral defenders},
  author={Qi, Jiahao and Gong, Zhiqiang and Liu, Xingyue and Chen, Chen and Zhong, Ping},
  journal={IEEE Transactions on Neural Networks and Learning Systems},
  volume={36},
  number={2},
  pages={3012--3026},
  year={2024},
  publisher={IEEE}
}

@article{dai2023instructblip,
  title={Instructblip: Towards general-purpose vision-language models with instruction tuning},
  author={Dai, Wenliang and Li, Junnan and Li, Dongxu and Tiong, Anthony and Zhao, Junqi and Wang, Weisheng and Li, Boyang and Fung, Pascale N and Hoi, Steven},
  journal={Advances in Neural Information Processing Systems},
  volume={36},
  pages={49250--49267},
  year={2023}
}

@article{achiam2023gpt,
  title={Gpt-4 technical report},
  author={Achiam, Josh and Adler, Steven and Agarwal, Sandhini and Ahmad, Lama and Akkaya, Ilge and Aleman, Florencia Leoni and Almeida, Diogo and Altenschmidt, Janko and Altman, Sam and Anadkat, Shyamal and others},
  journal={arXiv preprint arXiv:2303.08774},
  year={2023}
}

@inproceedings{jia2025foa,
  title={Adversarial Attacks against Closed-Source MLLMs via Feature Optimal Alignment},
  author={Jia, Xiaojun and Gao, Sensen and Qin, Simeng and Pang, Tianyu and Du, Chao and Huang, Yihao and Li, Xinfeng and Li, Yiming and Li, Bo and Liu, Yang},
  booktitle={Advances in Neural Information Processing Systems},
  year={2025}
}

@article{chai2026like,
  title={Like Human Rethinking: Contour Transformer AutoRegression for Referring Remote Sensing Interpretation},
  author={Chai, Jinming and Jiao, Licheng and Lu, Xiaoqiang and Li, Lingling and Liu, Fang and Sun, Long and Liu, Xu and Ma, Wenping and Li, Weibin},
  journal={IEEE Transactions on Pattern Analysis and Machine Intelligence},
  year={2026},
  publisher={IEEE}
}

@article{chen2024diffusion,
  title={Diffusion models for imperceptible and transferable adversarial attack},
  author={Chen, Jianqi and Chen, Hao and Chen, Keyan and Zhang, Yilan and Zou, Zhengxia and Shi, Zhenwei},
  journal={IEEE Transactions on Pattern Analysis and Machine Intelligence},
  volume={47},
  number={2},
  pages={961--977},
  year={2024},
  publisher={IEEE}
}

@article{zheng2025blackboxbench,
  title={Blackboxbench: A comprehensive benchmark of black-box adversarial attacks},
  author={Zheng, Meixi and Yan, Xuanchen and Zhu, Zihao and Chen, Hongrui and Wu, Baoyuan},
  journal={IEEE Transactions on Pattern Analysis and Machine Intelligence},
  year={2025},
  publisher={IEEE}
}

@article{li2025mini,
  title={Mini-gemini: Mining the potential of multi-modality vision language models},
  author={Li, Yanwei and Zhang, Yuechen and Wang, Chengyao and Zhong, Zhisheng and Chen, Yixin and Chu, Ruihang and Liu, Shaoteng and Jia, Jiaya},
  journal={IEEE Transactions on Pattern Analysis and Machine Intelligence},
  year={2025},
  publisher={IEEE}
}

@inproceedings{fu2024gptscore,
  title={Gptscore: Evaluate as you desire},
  author={Fu, Jinlan and Ng, See Kiong and Jiang, Zhengbao and Liu, Pengfei},
  booktitle={Proceedings of the North American Chapter of the Association for Computational Linguistics: Human Language Technologies},
  pages={6556--6576},
  year={2024}
}

@inproceedings{li2025mattack,
  title={A Frustratingly Simple Yet Highly Effective Attack Baseline: Over 90\% Success Rate Against the Strong Black-Box Models of GPT-4.5/4o/o1},
  author={Li, Zhaoyi and Zhao, Xiaohan and Wu, Dong-Dong and Cui, Jiacheng and Shen, Zhiqiang},
  booktitle={Advances in Neural Information Processing Systems},
  volume={38},
  year={2025}
}

@inproceedings{zhao2023mf-ii,
  title={On Evaluating Adversarial Robustness of Large Vision-Language Models},
  author={Zhao, Yunqing and Pang, Tianyu and Du, Chao and Yang, Xiao and Li, Chongxuan and Cheung, Ngai-Man and Lin, Min},
  booktitle={Advances in Neural Information Processing Systems},
  year={2023}
}

@inproceedings{cui2024robustness,
  title={On the robustness of large multimodal models against image adversarial attacks},
  author={Cui, Xuanming and Aparcedo, Alejandro and Jang, Young Kyun and Lim, Ser-Nam},
  booktitle={Proceedings of the IEEE/CVF Conference on Computer Vision and Pattern Recognition},
  pages={24625--24634},
  year={2024}
}

@article{guo2024efficient,
  title={Efficient Generation of Targeted and Transferable Adversarial Examples for Vision-Language Models via Diffusion Models},
  author={Guo, Qi and Pang, Shanmin and Jia, Xiaojun and Liu, Yang and Guo, Qing},
  journal={IEEE Transactions on Information Forensics and Security},
  year={2024}
}

@article{dong2023robust,
  title={How Robust is Google's Bard to Adversarial Image Attacks?},
  author={Dong, Yinpeng and Chen, Huanran and Chen, Jiawei and Fang, Zhengwei and Yang, Xiao and Zhang, Yichi and Tian, Yu and Su, Hang and Zhu, Jun},
  journal={arXiv preprint arXiv:2309.11751},
  year={2023}
}

@inproceedings{zhang2025anyattack,
  title={AnyAttack: Towards Large-Scale Self-Supervised Adversarial Attacks on Vision-Language Models},
  author={Zhang, Jiaming and Ye, Junhong and Ma, Xingjun and Li, Yige and Yang, Yunfan and Sang, Jitao and Yeung, Dit-Yan},
  booktitle={Proceedings of the IEEE/CVF Conference on Computer Vision and Pattern Recognition},
  year={2025}
}

@InProceedings{nie2026vattack,
  author    = {Nie, Sen and Zhang, Jie and Yan, Jianxin and Shan, Shiguang and Chen, Xilin},
  title     = {V-Attack: Targeting Disentangled Value Features for Controllable Adversarial Attacks on LVLMs},
  booktitle = {Proceedings of the IEEE/CVF Conference on Computer Vision and Pattern Recognition},
  year      = {2026},
  pages     = {42257--42267}
}

@article{wang2024qwen2,
  title={Qwen2-vl: Enhancing vision-language model's perception of the world at any resolution},
  author={Wang, Peng and Bai, Shuai and Tan, Sinan and Wang, Shijie and Fan, Zhihao and Bai, Jinze and Chen, Keqin and Liu, Xuejing and Wang, Jialin and Ge, Wenbin and others},
  journal={arXiv preprint arXiv:2409.12191},
  year={2024}
}

@article{liu2025survey,
  title={A survey of attacks on large vision--language models: Resources, advances, and future trends},
  author={Liu, Daizong and Yang, Mingyu and Qu, Xiaoye and Zhou, Pan and Cheng, Yu and Hu, Wei},
  journal={IEEE Transactions on Neural Networks and Learning Systems},
  year={2025},
  publisher={IEEE}
}

@inproceedings{liu2024safety,
  title={Safety of multimodal large language models on images and text},
  author={Liu, Xin and Zhu, Yichen and Lan, Yunshi and Yang, Chao and Qiao, Yu},
  booktitle={Proceedings of the International Joint Conference on Artificial Intelligence},
  pages={8151--8159},
  year={2024}
}

@InProceedings{Luu2024questioning,
    author    = {Luu, Duc-Tuan and Le, Viet-Tuan and Vo, Duc Minh},
    title     = {Questioning, Answering, and Captioning for Zero-Shot Detailed Image Caption},
    booktitle = {Proceedings of the Asian Conference on Computer Vision },
    year      = {2024},
    pages     = {242-259}
}

@inproceedings{lu2025benchmarking,
  title={Benchmarking Large Vision-Language Models via Directed Scene Graph for Comprehensive Image Captioning},
  author={Lu, Yiting and Wang, Xiaohan and Chen, Zhenfang and Xu, Xueming and Shen, Chunhua},
  booktitle={Proceedings of the IEEE/CVF Conference on Computer Vision and Pattern Recognition},
  year={2025}
}

@article{wang2024cross,
  title={Cross-Modal Retrieval: A Systematic Review of Methods and Future Directions},
  author={Wang, Tianshi and Li, Fengling and Zhu, Lei and Li, Jingjing and Zhang, Zheng and Shen, Heng Tao},
  journal={Proceedings of the IEEE},
  volume={112},
  number={11},
  pages={1716--1754},
  year={2024},
  publisher={IEEE}
}

@inproceedings{li2023blip,
  title     = {BLIP-2: Bootstrapping Language-Image Pre-training with Frozen Image Encoders and Large Language Models},
  author    = {Li, Junnan and Li, Dongxu and Savarese, Silvio and Hoi, Steven C. H.},
  booktitle = {Proceedings of the 40th International Conference on Machine Learning},
  pages     = {19730--19742},
  year      = {2023},
  volume    = {202},
  series    = {Proceedings of Machine Learning Research},
  publisher = {PMLR}
}

@inproceedings{zhu2024minigpt,
  title     = {MiniGPT-4: Enhancing Vision-Language Understanding with Advanced Large Language Models},
  author    = {Zhu, Deyao and Chen, Jun and Shen, Xiaoqian and Li, Xiang and Elhoseiny, Mohamed},
  booktitle = {Proceedings of the International Conference on Learning Representations},
  year      = {2024}
}

@article{yin2023vlattack,
  title={Vlattack: Multimodal adversarial attacks on vision-language tasks via pre-trained models},
  author={Yin, Ziyi and Ye, Muchao and Zhang, Tianrong and Du, Tianyu and Zhu, Jinguo and Liu, Han and Chen, Jinghui and Wang, Ting and Ma, Fenglong},
  journal={Advances in Neural Information Processing Systems},
  volume={36},
  pages={52936--52956},
  year={2023}
}

@article{hu2025rsgpt,
  title     = {RSGPT: A Remote Sensing Vision Language Model and Benchmark},
  author    = {Hu, Yuan and Yuan, Jianlong and Wen, Congcong and Lu, Xiaonan and Liu, Yu and Li, Xiang},
  journal   = {ISPRS Journal of Photogrammetry and Remote Sensing},
  volume    = {224},
  pages     = {272--286},
  year      = {2025},
  publisher = {Elsevier}
}

@inproceedings{kuckreja2024geochat,
  title     = {GeoChat: Grounded Large Vision-Language Model for Remote Sensing},
  author    = {Kuckreja, Kartik and Danish, Muhammad Sohail and Naseer, Muzammal and Das, Abhijit and Khan, Salman and Khan, Fahad Shahbaz},
  booktitle = {Proceedings of the IEEE/CVF Conference on Computer Vision and Pattern Recognition},
  pages     = {27831--27840},
  year      = {2024}
}

@inproceedings{muhtar2024lhrsbot,
  title     = {LHRS-Bot: Empowering Remote Sensing with {VGI}-Enhanced Large Multimodal Language Model},
  author    = {Muhtar, Dilxat and Li, Zhenshi and Gu, Feng and Zhang, Xueliang and Xiao, Pengfeng},
  booktitle = {Proceedings of the European Conference on Computer Vision},
  year      = {2024}
}

@inproceedings{irvin2025teochat,
  title     = {TEOChat: A Large Vision-Language Assistant for Temporal Earth Observation Data},
  author    = {Irvin, Jeremy and Liu, Emily and Chen, Joyce and Dormoy, Ines and Kim, Jinyoung and Khanna, Samar and Zheng, Zhuo and Ermon, Stefano},
  booktitle = {Proceedings of the International Conference on Learning Representations},
  year      = {2025}
}

@article{zhang2024vision,
  title={Vision-Language Models for Vision Tasks: A Survey},
  author={Zhang, Jingyi and Huang, Jiaxing and Jin, Sheng and Lu, Shijian},
  journal={IEEE Transactions on Pattern Analysis and Machine Intelligence},
  volume={46},
  number={8},
  pages={5625--5644},
  year={2024},
  publisher={IEEE}
}

@inproceedings{caron2021emerging,
  title={Emerging properties in self-supervised vision transformers},
  author={Caron, Mathilde and Touvron, Hugo and Misra, Ishan and J{\'e}gou, Herv{\'e} and Mairal, Julien and Bojanowski, Piotr and Joulin, Armand},
  booktitle = {Proceedings of the IEEE/CVF International Conference on Computer Vision},
  pages={9650--9660},
  year={2021}
}

@article{weng2025vision,
  title={Vision-language modeling meets remote sensing: Models, datasets, and perspectives},
  author={Weng, Xingxing and Pang, Chao and Xia, Gui-Song},
  journal={IEEE Geoscience and Remote Sensing Magazine},
  year={2025},
  publisher={IEEE}
}

@article{lin2025fedrsclip,
  title={Fedrsclip: Federated learning for remote sensing scene classification using vision-language models},
  author={Lin, Hui and Zhang, Chao and Hong, Danfeng and Dong, Kexin and Wen, Congcong},
  journal={IEEE Geoscience and Remote Sensing Magazine},
  year={2025},
  publisher={IEEE}
}

@article{chen2026integration,
  title={Integration of large vision language models for efficient post-disaster damage assessment and reporting},
  author={Chen, Zhaohui and Asadi Shamsabadi, Elyas and Jiang, Sheng and Shen, Luming and Dias-da-Costa, Daniel},
  journal={Nature Communications},
  year={2026},
  publisher={Nature Publishing Group UK London}
}

@article{li2019deep,
  title={Deep neural network for remote-sensing image interpretation: Status and perspectives},
  author={Li, Jiayi and Huang, Xin and Gong, Jianya},
  journal={National Science Review},
  volume={6},
  number={6},
  pages={1082--1086},
  year={2019},
  publisher={Oxford University Press}
}

@article{gemini2023gemini,
  title   = {Gemini: A Family of Highly Capable Multimodal Models},
  author  = {{Gemini Team}},
  journal = {arXiv preprint arXiv:2312.11805},
  year    = {2023},
}

@techreport{openai2025gpt5,
  title       = {GPT-5 System Card},
  author      = {{OpenAI}},
  institution = {OpenAI},
  year        = {2025},

}

@techreport{anthropic2025claude4,
  title       = {System Card: Claude Opus 4 \& Claude Sonnet 4},
  author      = {{Anthropic}},
  institution = {Anthropic},
  year        = {2025},
}

@article{liu2024remoteclip,
  title   = {RemoteCLIP: A Vision Language Foundation Model for Remote Sensing},
  author  = {Liu, Fan and Chen, Delong and Guan, Zhangqingyun and Zhou, Xiaocong and Zhu, Jiale and Ye, Qiaolin and Fu, Liyong and Zhou, Jun},
  journal = {IEEE Transactions on Geoscience and Remote Sensing},
  volume  = {62},
  pages   = {1--16},
  year    = {2024},
  doi     = {10.1109/TGRS.2024.3390838}
}

@article{zhan2025skyeyegpt,
  title   = {SkyEyeGPT: Unifying Remote Sensing Vision-Language Tasks via Instruction Tuning with Large Language Model},
  author  = {Zhan, Yang and Xiong, Zhitong and Yuan, Yuan},
  journal = {ISPRS Journal of Photogrammetry and Remote Sensing},
  volume  = {221},
  pages   = {64--77},
  year    = {2025}
}

@article{zhang2024earthgpt,
  title={Earthgpt: A universal multi-modal large language model for multi-sensor image comprehension in remote sensing domain},
  author={Zhang, Wei and Cai, Miaoxin and Zhang, Tong and Zhuang, Yin and Mao, Xuerui},
  journal={IEEE Transactions on Geoscience and Remote Sensing},
  year={2024},
  publisher={IEEE}
}

@inproceedings{liu2017delving,
    title={Delving into transferable adversarial examples and black-box attacks},
    author={Liu, Yanpei and Chen, Xinyun and Liu, Chang and Song, Dawn},
    booktitle={Proceedings of the International Conference on Learning Representations},
    year={2017}
}

@inproceedings{radford2021learning,
  title={Learning transferable visual models from natural language supervision},
  author={Radford, Alec and Kim, Jong Wook and Hallacy, Chris and Ramesh, Aditya and Goh, Gabriel and Agarwal, Sandhini and Sastry, Girish and Askell, Amanda and Mishkin, Pamela and Clark, Jack and others},
  booktitle={Proceedings of the International Conference on Machine Learning},
  pages={8748--8763},
  year={2021}
}

@inproceedings{li2026mpcattack,
  title     = {Multi-Paradigm Collaborative Adversarial Attack Against Multi-Modal Large Language Models},
  author    = {Li, Yuanbo and Xu, Tianyang and Hu, Cong and Zhou, Tao and Wu, Xiao-Jun and Kittler, Josef},
  booktitle = {Proceedings of the IEEE/CVF Conference on Computer Vision and Pattern Recognition},
  year      = {2026}
}

@inproceedings{mei2026veattack,
  title     = {{VEAttack}: Downstream-Agnostic Vision Encoder Attack against Large Vision Language Models},
  author    = {Mei, Hefei and Wang, Zirui and You, Shen and Dong, Minjing and Xu, Chang},
  booktitle = {International Conference on Learning Representations},
  year      = {2026}
}

@article{vaswani2017attention,
  title={Attention is all you need},
  author={Vaswani, Ashish and Shazeer, Noam and Parmar, Niki and Uszkoreit, Jakob and Jones, Llion and Gomez, Aidan N and Kaiser, {\L}ukasz and Polosukhin, Illia},
  journal={Advances in Neural Information Processing Systems},
  volume={30},
  year={2017}
}

@inproceedings{madry2018towards,
    title={Towards deep learning models resistant to adversarial attack},
    author={Madry, Aleksander and Makelov, Aleksandar and Schmidt, Ludwig and Tsipras, Dimitris and Vladu, Adrian},
    booktitle={Proceedings of the International Conference on Learning Representations},
    year={2018}
}

@inproceedings{dong2018boosting,
    title={Boosting adversarial attacks with momentum},
    author={Dong, Yinpeng and Liao, Fangzhou and Pang, Tianyu and Su, Hang and Zhu, Jun and Hu, Xiaolin and Li, Jianguo},
    booktitle={Proceedings of the IEEE Conference on Computer Vision and Pattern Recognition},
    pages={9185--9193},
    year={2018}
}

@inproceedings{kurakin2017adversarial,
    title={Adversarial examples in the physical world},
    author={Kurakin, Alexey and Goodfellow, Ian J. and Bengio, Samy},
    booktitle={Proceedings of the International Conference on Learning Representations Workshop},
    year={2017}
}

@inproceedings{kurakin2016adversarial,
  title={Adversarial machine learning at scale},
  author={Kurakin, Alexey and Goodfellow, Ian and Bengio, Samy},
  booktitle={Proceedings of International Conference on Learning Representations},
  year={2016}
}

@article{xu2020assessing,
  title={Assessing the threat of adversarial examples on deep neural networks for remote sensing scene classification: Attacks and defenses},
  author={Xu, Yonghao and Du, Bo and Zhang, Liangpei},
  journal={IEEE Transactions on Geoscience and Remote Sensing},
  volume={59},
  number={2},
  pages={1604--1617},
  year={2020},
  publisher={IEEE}
}

@article{chen2021empirical,
  title   = {An Empirical Study of Adversarial Examples on Remote Sensing Image Scene Classification},
  author  = {Chen, Li and Xu, Zewei and Li, Qi and Peng, Jing and Wang, Shaocong and Li, Haifeng},
  journal = {IEEE Transactions on Geoscience and Remote Sensing},
  volume  = {59},
  number  = {9},
  pages   = {7419--7433},
  year    = {2021}
}

@article{xu2022universal,
  title={Universal adversarial examples in remote sensing: Methodology and benchmark},
  author={Xu, Yonghao and Ghamisi, Pedram},
  journal={IEEE Transactions on Geoscience and Remote Sensing},
  volume={60},
  pages={1--15},
  year={2022},
  publisher={IEEE}
}

@article{fu2024sfcot,
    title={Transferable Adversarial Attacks for Remote Sensing Object Recognition via Spatial-Frequency Co-Transformation},
    author={Fu, Yimin and Liu, Zhunga and Lyu, Jialin},
    journal={IEEE Transactions on Geoscience and Remote Sensing},
    year={2024},
    publisher={IEEE}
}

@article{shi2021hyperspectral,
  title   = {Hyperspectral Image Classification With Adversarial Attack},
  author  = {Shi, Cheng and Dang, Yenan and Fang, Leyuan and Lv, Zhiyong and Zhao, Minghua},
  journal = {IEEE Geoscience and Remote Sensing Letters},
  volume  = {19},
  pages   = {1--5},
  year    = {2022}
}

@article{bai2024stealthy,
  title   = {Stealthy Adversarial Examples for Semantic Segmentation in Remote Sensing},
  author  = {Bai, Tao and Cao, Yiming and Xu, Yonghao and Wen, Bihan},
  journal = {IEEE Transactions on Geoscience and Remote Sensing},
  volume  = {62},
  pages   = {1--17},
  year    = {2024},
  doi     = {10.1109/TGRS.2024.3377009}
}

@article{su2023reconstruction,
  title   = {Reconstruction-Assisted and Distance-Optimized Adversarial Training: A Defense Framework for Remote Sensing Scene Classification},
  author  = {Su, Yuru and Zhang, Ge and Mei, Shaohui and Lian, Jiawei and Wang, Ye and Wan, Shuai},
  journal = {IEEE Transactions on Geoscience and Remote Sensing},
  volume  = {61},
  pages   = {1--13},
  year    = {2023},
}

@article{yu2024diffusion,
  title   = {Universal Adversarial Defense in Remote Sensing Based on Pre-Trained Denoising Diffusion Models},
  author  = {Yu, Weikang and Xu, Yonghao and Ghamisi, Pedram},
  journal = {International Journal of Applied Earth Observation and Geoinformation},
  volume  = {133},
  pages   = {104131},
  year    = {2024},
}

@article{liu2026scattering,
  title   = {Scattering-Guided Class-Irrelevant Filtering for Adversarially Robust SAR Automatic Target Recognition},
  author  = {Liu, Zhunga and Lyu, Jialin and Fu, Yimin},
  journal = {Signal Processing},
  volume  = {239},
  pages   = {110273},
  year    = {2026},
}

@article{chang2024survey,
  title={A survey on evaluation of large language models},
  author={Chang, Yupeng and Wang, Xu and Wang, Jindong and Wu, Yuan and Yang, Linyi and Zhu, Kaijie and Chen, Hao and Yi, Xiaoyuan and Wang, Cunxiang and Wang, Yidong and others},
  journal={ACM Transactions on Intelligent Systems and Technology},
  volume={15},
  number={3},
  pages={1--45},
  year={2024},
  publisher={ACM New York, NY}
}

@article{peng2022scattering,
  title={Scattering model guided adversarial examples for SAR target recognition: Attack and defense},
  author={Peng, Bowen and Peng, Bo and Zhou, Jie and Xie, Jianyue and Liu, Li},
  journal={IEEE Transactions on Geoscience and Remote Sensing},
  volume={60},
  pages={1--17},
  year={2022},
  publisher={IEEE}
}

@inproceedings{pathak2024model,
  title     = {Model Agnostic Defense Against Adversarial Patch Attacks on Object Detection in Unmanned Aerial Vehicles},
  author    = {Pathak, Saurabh and Shrestha, Samridha and AlMahmoud, Abdelrahman},
  booktitle = {Proceedings of the 2024 IEEE/RSJ International Conference on Intelligent Robots and Systems},
  pages     = {2586--2593},
  year      = {2024},
}

@article{xu2021sacnet,
  title   = {Self-Attention Context Network: Addressing the Threat of Adversarial Attacks for Hyperspectral Image Classification},
  author  = {Xu, Yonghao and Du, Bo and Zhang, Liangpei},
  journal = {IEEE Transactions on Image Processing},
  volume  = {30},
  pages   = {8671--8685},
  year    = {2021},
}

@article{fu2026transferability,
  title={Transferability Reinforcement of Adversarial Attacks for Remote Sensing Image Classification via Hierarchical Transformation Composition},
  author={Fu, Yimin and Bai, Yuefeng and Lyu, Jialin and Pan, Baicheng and Liu, Zhunga and Ng, Michael K},
  journal={IEEE Transactions on Geoscience and Remote Sensing},
  year={2026},
  publisher={IEEE}
}

@inproceedings{yang2010ucm,
	title     = {Bag-of-Visual-Words and Spatial Extensions for Land-Use Classification},
	author    = {Yang, Yi and Newsam, Shawn},
	booktitle = {Proceedings of the 18th ACM SIGSPATIAL International Conference on Advances in Geographic Information Systems},
	pages     = {270--279},
	year      = {2010}
}

@article{zhao2015siri,
    title={Dirichlet-Derived Multiple Topic Scene Classification Model for High Spatial Resolution Remote Sensing Imagery},
    author={{Zhao, Bei and Zhong, Yanfei and Xia, Guisong and Zhang, Liangpei}},
    journal={IEEE Transactions on Geoscience and Remote Sensing},
    volume={54},
    number={4},
    pages={2108--2123},
    year={2015},
    publisher={IEEE}
}

@article{xia2017aid,
  title={AID: A Benchmark Data Set for Performance Evaluation of Aerial Scene Classification},
  author={Xia, Gui-Song and Hu, Jingwen and Hu, Fan and Shi, Baoguang and Bai, Xiang and Zhong, Yanfei and Zhang, Liangpei},
  journal={IEEE Transactions on Geoscience and Remote Sensing},
  volume={55},
  number={7},
  pages={3965--3981},
  year={2017},
  publisher={IEEE},
  doi={10.1109/TGRS.2017.2685945}
}

@article{bai2025qwen25vl,
  title   = {Qwen2.5-VL Technical Report},
  author  = {Bai, Shuai and Chen, Keqin and Liu, Xuejing and
             Wang, Jialin and Ge, Wenbin and Song, Sibo and
             Dang, Kai and Wang, Peng and Wang, Shijie and
             Tang, Jun and others},
  journal = {arXiv preprint arXiv:2502.13923},
  year    = {2025}
}

@article{comanici2025gemini25,
  title={Gemini 2.5: Pushing the Frontier with Advanced Reasoning, Multimodality, Long Context, and Next Generation Agentic Capabilities},
  author={Comanici, Gheorghe and Bieber, Eric and Schaekermann, Mike and others},
  journal={arXiv preprint arXiv:2507.06261},
  year={2025}
}

\end{document}